\documentclass{article}
\usepackage{spconf}
\usepackage{times}
\usepackage{epsfig}
\usepackage{graphicx}
\usepackage{amsmath}
\usepackage{amssymb}
\usepackage[font=small,skip=4pt]{caption}
\usepackage{subcaption}
\usepackage{overpic}
\usepackage{xcolor}
\usepackage[french,english]{babel}
\newcommand{\etal}{\textit{et al. }}
\newcommand{\eg}{\textit{e.g.}}

\title{Photo style transfer with consistency losses}
\name{Xu Yao$^{1, 2}\sthanks{Work done when Xu Yao was a last-year student at Mines ParisTech.}$ \qquad Gilles Puy$^{1}$ \qquad Patrick P\'erez$^{3}$}

\address{
$^{1}$Technicolor R\&I, 975 avenue des Champs Blancs, 35576 Cesson-S\'evign\'e, France \\
$^{2}$Telecom ParisTech, 46 Rue Barrault, 75013 Paris, France \\
$^{3}$Valeo.ai, 15 rue de la Baume, 75008 Paris, France}

\begin{document}
%
\maketitle
%
\begin{abstract}
We address the problem of 
style transfer between two photos and propose a new way to preserve photorealism. Using the single pair of photos available as input, we train a pair of deep convolution networks (convnets), each of which transfers the style of one photo to the other. To enforce photorealism, we introduce a content preserving mechanism by combining a cycle-consistency loss with a self-consistency loss. Experimental results show that this method does not suffer from typical artifacts observed in methods working in the same settings \cite{luan17,li2018}. 
We then further analyze some properties of these trained convnets. First, we notice that they can be used to stylize other unseen images with same known style.
Second, we show that retraining only a small subset of the network parameters can be sufficient to adapt these convnets to new styles.
\end{abstract}

%

\section{Introduction}
\label{sec:intro}
Image style transfer has been investigated for many years \cite{hertzmann2001image, efros2001image, reinhard2001color}. In the case where one wants to make a photo look like a painting, several innovative approaches using deep convnets have been proposed recently \cite{gatys16, johnson2016perceptual, ulyanov2016texture, dumoulin17, ghiasi17, huang17, li17}. 
However, these methods fail for style transfer between photos as they generate strong artifacts and the results lack photorealism. This issue can be partly solved by applying a structure-preserving filter / regularization \cite{luan17, li2018, puy2018} based on the ``matting Laplacian'' \cite{levin08}. But since this filter tries to generate an eye-pleasing image from an intermediate result with heavy distortions, it sometimes fails to recover all the structures visible in the input photo or to suppress stylization artifacts. 
In this work, we propose an alternative method for photo style transfer.

The recent work of Ulyanov \etal\cite{ulyanov2017dip} shows that the sole structure of a deep convnet is able to capture low-level image statistics before any learning and, thus, can be used as an image prior. 
Given a deteriorated image, they show that optimizing the parameters of a randomly-initialized convnet, using only this image for ``training'', is sufficient to improve its quality. The authors demonstrate the efficiency of this technique for, \eg, denoising, up-sampling and inpainting. Inspired by this work, we propose to train two networks for photo stylization using only the available pair of photos. All along the training, the networks are forced to preserve the structure of the input photos via cycle- and self-consistency mechanisms \cite{zhu2017}. This approach successfully prevents structure distortions and generates satisfying photo stylization. This constitutes our main contribution. We compare our results to those obtained with state-of-the-art methods in Section~\ref{sec:results}. We then explore some properties of the trained convnets. 
First, even though these networks are trained using only a single pair of photos, we show that each network can transfer the learnt style to natural images not viewed at training time. The results are comparable to those obtained by training two new convnets using each new image and the original photo of the style. We note however that the images to be stylized need to have a similar semantic content to the original photo of the style.
Second, to apply a new style, we show that only a small subset of the network parameters needs to be retrained (with however a good choice of the image pair used to pre-train the other parameters). 

\textbf{Related works.} Optimization-based Neural Style Transfer \cite{gatys16} has attracted a wide interest as this is the first method to perform artistic style transfer using features extracted from a deep convnet. This method being slow, follow-up works \cite{ulyanov2016texture, johnson2016perceptual, dumoulin17, ghiasi17, huang17, li17} proposed 
fast artistic stylization by training feed-forward networks to estimate the solution of \cite{gatys16} on a large collection of images.
An improvement of the neural style transfer method was proposed by \cite{luan17} to address photo stylization, with faster methods detailed in \cite{li2018, puy2018}. 
In this work, we first concentrate on improving \cite{luan17} with the construction of a new photo prior. This prior is constructed using two deep convnets that are trained from a random initialization using two images as inputs. Once trained each network can be used to stylize rapidly new images towards the style of one of the images used for training. Applying a new style needs however a partial retraining.
Another related image transformation task is domain adaptation. Each style could be considered as a domain and deep convnets can be trained to transform images from a source domain to a target domain \cite{pix2pix2016, zhu2017, wang2018pix2pixHD}. In this work, we use the cycle-consistency mechanism widely used for unpaired domain adaptation. Note however that we do not use any adversarial loss and that a domain/style in our case is not made of a collection but of a single image.

\section{Photo Style Transfer Network}
\label{sec:prior}

Our goal is to transfer the style of an image $\mathbf{x}_a$ to another image $\mathbf{x}_b$. 
We denote this stylized image by $\mathbf{x}_{b \rightarrow a}$. Following \cite{luan17}, $\mathbf{x}_{b \rightarrow a}$ is defined as a solution of 
\begin{equation}
\label{eq:style_loss}
\arg\min_{\mathbf{x}} \lambda_c \, \mathcal{L}_{c}(\mathbf{x}, \mathbf{x}_b) + \lambda_s \, \mathcal{L}_{s}(\mathbf{x}, \mathbf{x}_a) + \lambda_{\mathsf{L}} \, \mathcal{L}_{\mathsf{L}}(\mathbf{x}),
\end{equation}
where $\lambda_c, \lambda_s, \lambda_{\mathsf{L}} > 0$. The content loss $\mathcal{L}_{c}$ permits us to retain the content of $\mathbf{x}_b$ in $\mathbf{x}_{b \rightarrow a}$, while the style loss $\mathcal{L}_{s}$ aims at transferring the style of $\mathbf{x}_a$. 
The content loss is defined as the Euclidean distance between VGG-19 features \cite{simonyan14} of $\mathbf{x}_b$ and $\mathbf{x}$. The style loss is defined as the Euclidean distance between Gram matrices of VGG-19 features of $\mathbf{x}_a$ and $\mathbf{x}$. Finally, $\mathcal{L}_{\mathsf{L}}$ is a regularization term favoring photorealism and built using the matting Laplacian $\mathsf{L}$ \cite{levin08}. Due to space constraint, we let the reader refer to \cite{luan17} for more details about these losses.

Although visually pleasing at the first glance, the photo stylization results obtained with the use of the matting Laplacian $\mathsf{L}$ present several undesirable artifacts (see Section~\ref{sec:results}). To avoid these disadvantages, we exploit the fact that both $\mathbf{x}_a$ and $\mathbf{x}_b$ are photorealistic to build a new content loss that preserves photorealism as much as possible. 

\subsection{Content preserving mechanism}

\begin{figure}
\begin{subfigure}[b]{0.49\linewidth}
\begin{flushleft}
\includegraphics[width=0.95\linewidth]{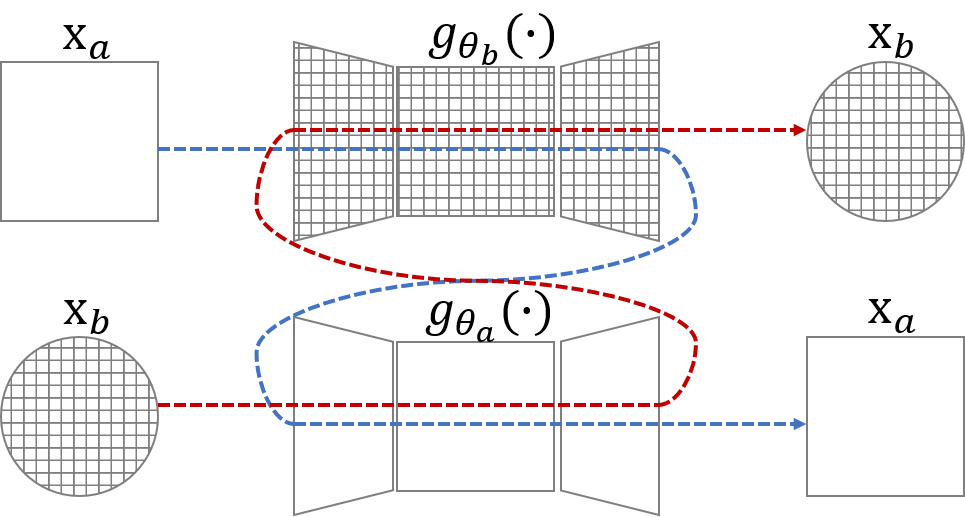}
\caption{Cycle-consistency}
\end{flushleft}
\end{subfigure}
\begin{subfigure}[b]{0.49\linewidth}
\begin{flushright}
\includegraphics[width=0.95\linewidth]{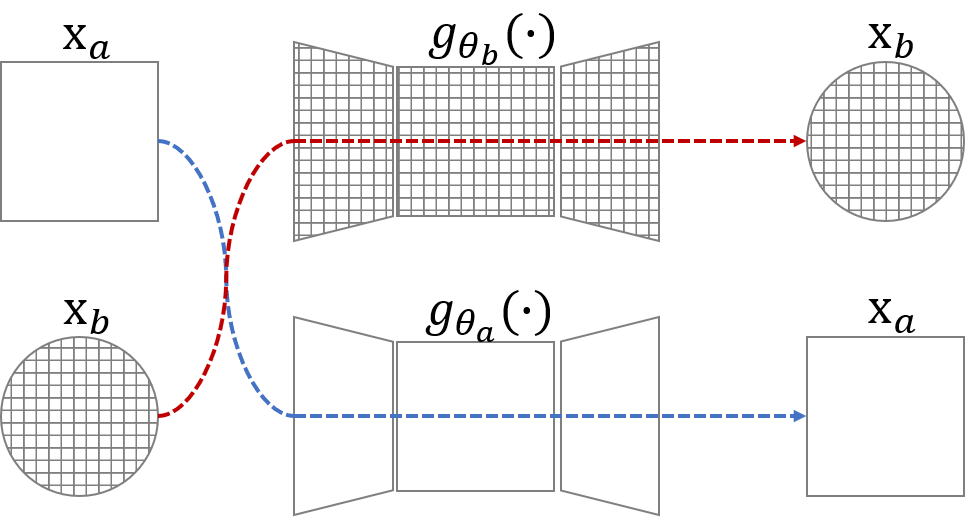}
\caption{Self-consistency}
\end{flushright}
\end{subfigure}
\vspace{10pt}
   \caption{The network $g_{\theta_a}(\cdot)$ transfers the style of the image $\mathbf{x}_a$ to its input, while $g_{\theta_b}(.)$ transfers the style of $\mathbf{x}_b$. Cycle-consistency is the fact that $g_{\theta_b}(g_{\theta_a}(\mathbf{x}_b)) \approx \mathbf{x}_b$ and $g_{\theta_a}(g_{\theta_b}(\mathbf{x}_a)) \approx \mathbf{x}_a$, while self-consistency is the fact that $g_{\theta_b}(\mathbf{x}_b) \approx \mathbf{x}_b$ and $g_{\theta_a}(\mathbf{x}_a) \approx \mathbf{x}_a$.
   }
\label{twonetwork}
\end{figure}

First, we remove the loss $\mathcal{L}_{\mathsf{L}}$ involving the matting Laplacian in \eqref{eq:style_loss}, or, equivalently, set $\lambda_{\mathsf{L}}=0$.

Second, taking inspiration from \cite{ulyanov2017dip}, we propose to use the constraint $\mathbf{x} = g_{\theta_a}(\mathbf{x}_b)$ in \eqref{eq:style_loss}, where $g_{\theta_a}(\cdot)$ is a deep network with parameters $\theta_a$ that performs stylization toward the style of $\mathbf{x}_a$. Therefore, instead of directly minimizing the loss on $\mathbf{x}$, the minimization is conducted over $\theta_a$.

Third, we borrow from \cite{zhu2017} the idea of cycle consistency to construct a new content loss preserving input structures. Let $g_{\theta_b}(\cdot)$ be a deep network with parameters $\theta_b$ that performs stylization toward the style of $\mathbf{x}_b$. If we feed the stylized image $\mathbf{x}_{b \rightarrow a} = g_{\theta_a}(\mathbf{x}_b)$ into $g_{\theta_b}(\cdot)$, we expect to recover the original input image $\mathbf{x}_b$. Similarly, we expect $g_{\theta_a}(g_{\theta_b}(\mathbf{x}_a)) = \mathbf{x}_a$ (Fig.~\ref{twonetwork}a). The first part of our new content loss reads
\begin{align}
\label{eq:cycle_consistency}
\tilde{\mathcal{L}}_{1} (\theta_a, \theta_b) 
= \mathcal{L}_{c}(g_{\theta_b}(g_{\theta_a}({\mathbf{x}_b})),&\, {\mathbf{x}_b}) \nonumber\\
+ &\mathcal{L}_{c}(g_{\theta_a}(g_{\theta_b}({\mathbf{x}_a})), {\mathbf{x}_a}),
\end{align}
where we recall that $ \mathcal{L}_{c}$ is a perceptual loss constructed using VGG-19 features. As in the recent work on image inpainting \cite{huy2018}, we use layers {\tt conv1$\_$1}, {\tt conv2$\_$1} and {\tt conv3$\_$1} of VGG-19.
\begin{figure}
\scriptsize
\begin{subfigure}[b]{0.24\linewidth}
\centering
Input
\includegraphics[width=\linewidth, height=0.62\linewidth]{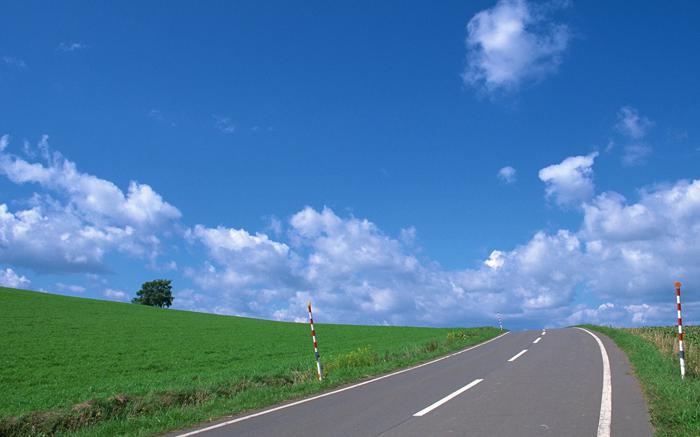}

\vspace{0.05cm}
\includegraphics[width=\linewidth, height=0.62\linewidth]{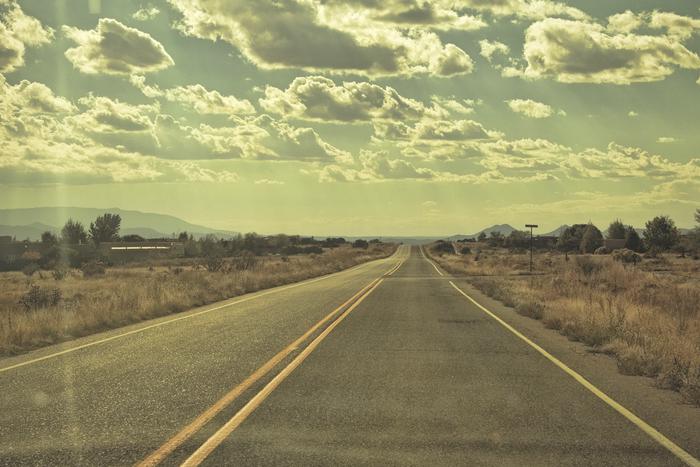}
\end{subfigure}
\begin{subfigure}[b]{0.24\linewidth}
\centering
Cycle-consistency
\includegraphics[width=\linewidth, height=0.62\linewidth]{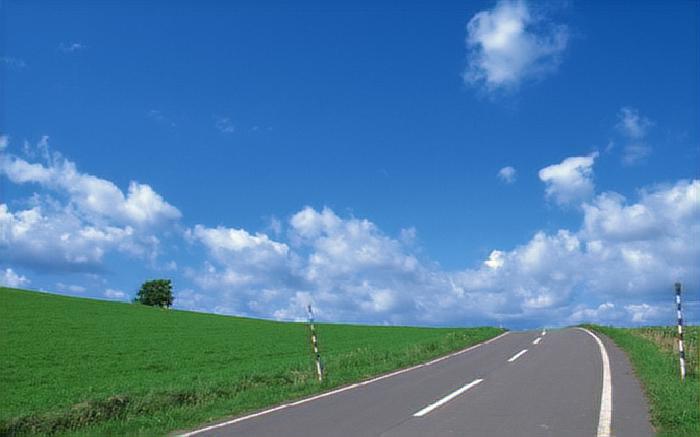}

\vspace{0.05cm}
\includegraphics[width=\linewidth, height=0.62\linewidth]{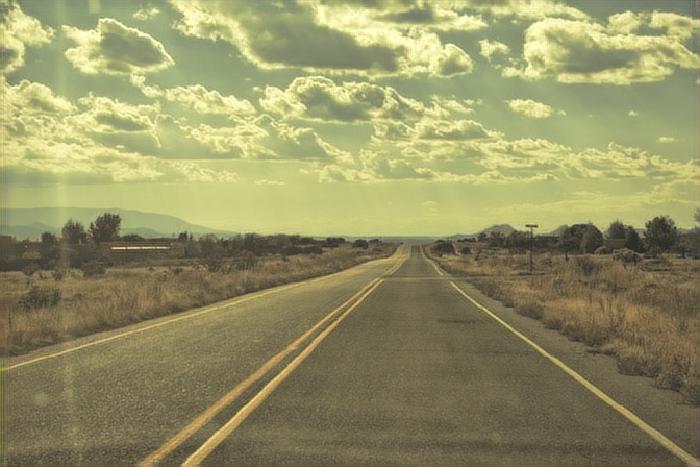}
\end{subfigure}
\begin{subfigure}[b]{0.24\linewidth}
\centering
Self-consistency
\includegraphics[width=\linewidth, height=0.62\linewidth]{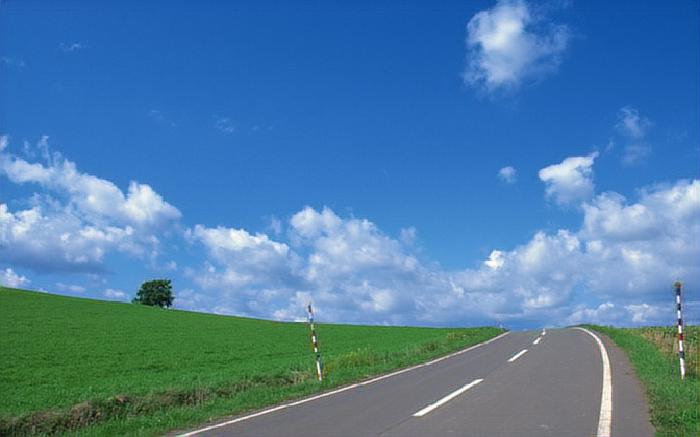}

\vspace{0.05cm}
\includegraphics[width=\linewidth, height=0.62\linewidth]{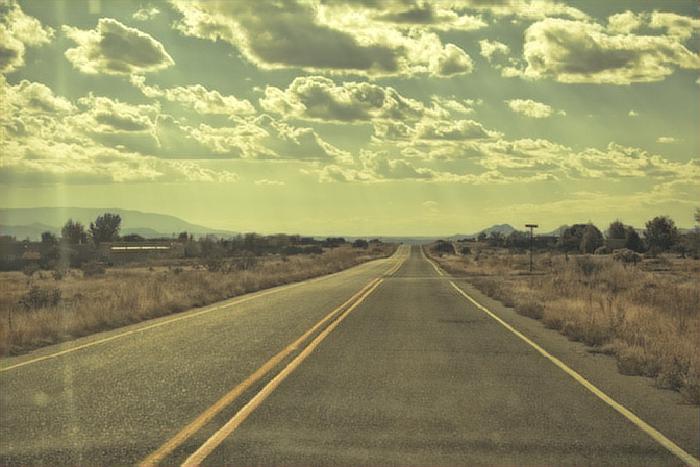}
\end{subfigure}
\begin{subfigure}[b]{0.24\linewidth}
\centering
Stylization
\includegraphics[width=\linewidth, height=0.62\linewidth]{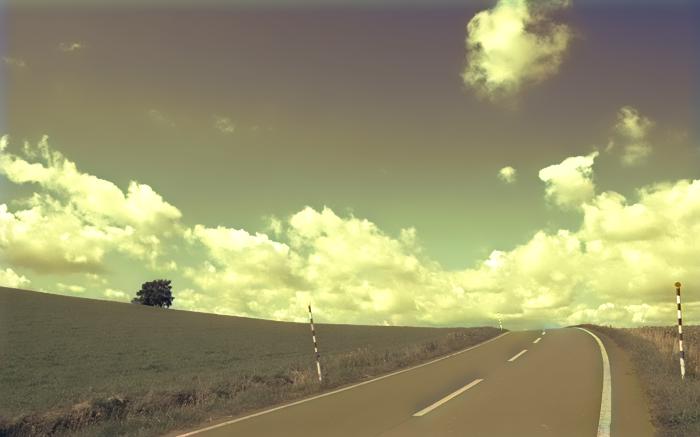}

\vspace{0.05cm}
\includegraphics[width=\linewidth, height=0.62\linewidth]{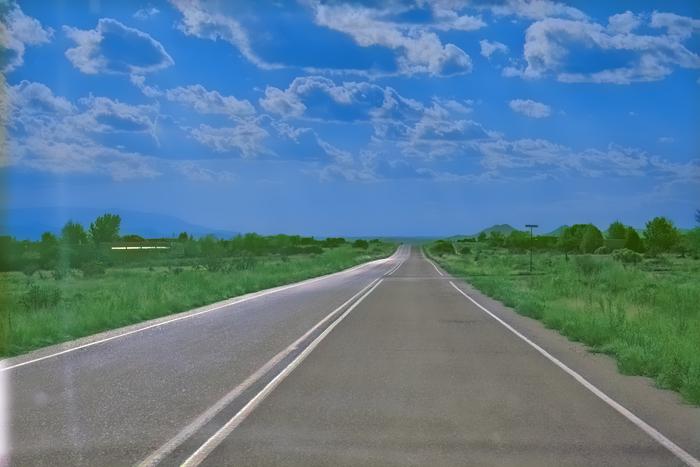}
\end{subfigure}
\vspace{10pt}
   \caption{
Given a pair of input photos, the proposed cycle/self-consistencies yield images that are almost identical to the inputs while the stylization renders each photo in the style of the other.
   }
\label{cycle-self-consistency}
\end{figure}
 
Fourth, given the input image $\mathbf{x}_a$, the stylization network ${g_{\theta_a}(\cdot)}$ should be able to preserve $\mathbf{x}_a$: $\mathbf{x}_a  = g_{\theta_a}(\mathbf{x}_a)$. Similarly, $\mathbf{x}_b = g_{\theta_b}(\mathbf{x}_b)$. We name this mechanism self-consistency (Fig.~\ref{twonetwork}b). This yields the second part of our new content loss:
\begin{eqnarray}
\label{eq:self_consistency}
\tilde{\mathcal{L}}_{2} (\theta_a, \theta_b) 
= \mathcal{L}_{c}(g_{\theta_a}({\mathbf{x}_a}), {\mathbf{x}_a}) 
+ \mathcal{L}_{c}(g_{\theta_b}({\mathbf{x}_b}), {\mathbf{x}_b}).
\end{eqnarray}
We remarked that using the cycle-consistency loss exclusively as content loss was not sufficient to preserve input structures. 
Using jointly the cycle-consistency and self-consistency losses improved the quality of our results. We show in Fig.~\ref{cycle-self-consistency} that the cyle- and self-consistencies are well respected by the pair of trained networks.

Finally, the stylization is controlled by the style loss $\mathcal{L}_{s}$, calculated for both styles:
\begin{eqnarray}
\label{eq:joint_style_loss}
\tilde{\mathcal{L}}_{s}(\theta_a, \theta_b) 
= \mathcal{L}_{s}(g_{\theta_a}({\mathbf{x}_b}), {\mathbf{x}_a}) 
+ \mathcal{L}_{s}(g_{\theta_b}({\mathbf{x}_a}), {\mathbf{x}_b}),
\end{eqnarray}
where we recall that $\mathcal{L}_{s}$ is defined by the distance between Gram matrices of VGG-19 features. We use layers {\tt conv1$\_$1}, {\tt conv2$\_$1} and {\tt conv3$\_$1}.

In total, we replace \eqref{eq:style_loss} 
by minimizing w.r.t. networks' parameters the complete loss: 
\begin{eqnarray}
\label{eq:equation5}
\tilde{\mathcal{L}}(\theta_a, \theta_b)  = 
\lambda_c  \left[\tilde{\mathcal{L}}_{1} (\theta_a, \theta_b) + \tilde{\mathcal{L}}_{2} (\theta_a, \theta_b) \right] 
+ \lambda_s \, \tilde{\mathcal{L}}_{s}(\theta_a, \theta_b).
\end{eqnarray}

In \cite{zhu2017}, the cycle-consistency loss is used in combination with an adversarial loss, which also contributes to the photorealism of their result. In our case, no adversarial loss is used. Let us also re-emphasize that, unlike in \cite{zhu2017} and related works, $g_{\theta_a}(\cdot)$ and $g_{\theta_b}(\cdot)$ are trained using only two images: $\mathbf{x}_a$ and $\mathbf{x}_b$. Our method is also inspired by the work of \cite{ulyanov2017dip}, where one network is trained using one image (and a fixed random input) to perform, \eg, denoising, up-sampling and inpainting. However, our trained networks can be applied to images not viewed at training time while the network trained in \cite{ulyanov2017dip} remains specific to the image used for training.

\subsection{Network architecture}

\begin{figure}
\begin{center}
\includegraphics[width=\linewidth]{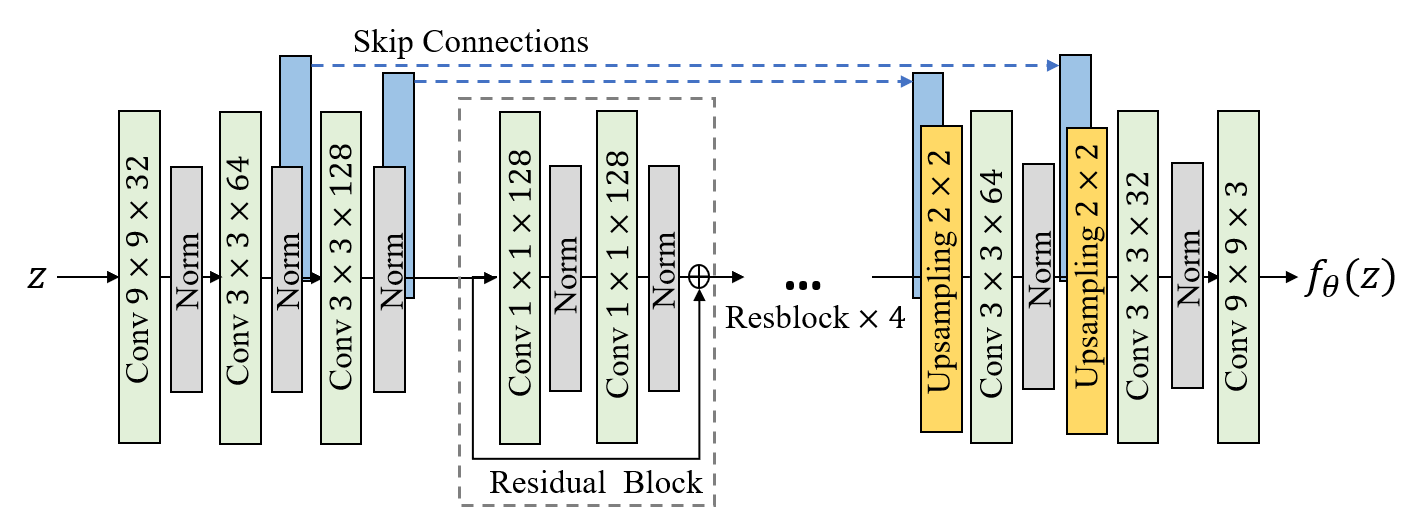}
\vspace{1pt}
\caption{Our networks consists of 16 convolution layers, of which 10 are in the residual blocks. Instance normalization is added after each layer except the last one. The 2nd and 3rd layers have a stride of 2. Two skip connections are added over the residual blocks.}
\label{arch}
\end{center}
\end{figure}

We use a network architecture which has been proved effective by earlier works on artistic style transfer \cite{johnson2016perceptual, dumoulin17} and domain adaptation \cite{zhu2017}. The structure of the network is presented in Fig.~\ref{arch}. It is similar to that of \cite{dumoulin17} with two differences. First, inspired by U-net \cite{unet2015}, we introduce skip connections between the second and penultimate layers, as well as between the third and ante-penultimate layers, with the aim of better preserving the structure of the input image. 

Second, we also reduce the kernel size in the residual blocks \cite{he_resnet16} from ${3 \times 3}$ to ${1 \times 1}$ as, in our case, reducing the number of parameters had no impact on the quality of generated results while accelerating the training. 

\subsection{Implementation details}
\label{Implemention Details}

As in \cite{luan17, li2018, puy2018}, we match the style of similar semantic regions (sky, building, lake, \emph{etc.}) between two images by using semantic segmentation masks.
We allocate one stylization network for each semantic region to prevent style mixing. Hence, for a pair of photos each with $n$ corresponding semantic regions, we train $2n$ networks. 
Training is performed jointly for up to 8 semantic regions. 

The style given to an image in the networks $g(\cdot)$ can be controlled by the instance normalization parameters as shown in \cite{dumoulin17, huang17}. 
In order to reduce the number of trainable parameters, the two networks share the same convolutional layers but have different instance normalization layers. 

The full objective function \eqref{eq:equation5} contains six sub-losses\footnote{Two in \eqref{eq:cycle_consistency}, two in \eqref{eq:self_consistency}, two in \eqref{eq:joint_style_loss}.}, each involving
the VGG-19 network. 
To reduce memory footprint, we randomly draw one of the sub-losses at each iteration and apply a gradient step using this sub-loss only. The sub-losses are drawn using a uniform distribution and without replacement. This ensures that all the sub-losses are selected once every six iterations. For a pair of photos of size $700\times400$, each with a single semantic region, the training process takes about 15 minutes on a NVIDIA Tesla P100 GPU. Our method was implemented using PyTorch \cite{paszke2017automatic}.

For a fair comparison with existing methods, we post-process the obtained images using the same filter as in \cite{luan17, li2018, puy2018}. This post-processing permits ones to enhance fine details and global photorealism.

\section{Results}
\label{sec:results}

\subsection{Comparison with state-of-the-art methods}
\begin{figure}
\begin{center}
\captionsetup[subfigure]{font=scriptsize}
\small

\hspace{5pt}
\begin{subfigure}[b]{0.185\linewidth}
\centering
Input
\begin{overpic}[width=\linewidth, height=0.62\linewidth]{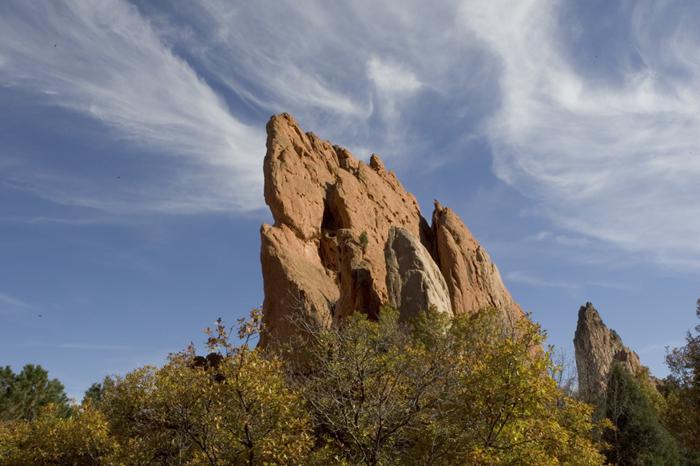}
\put(-2,38){\color{white}%
	\frame{\includegraphics[width=0.4\linewidth, height=0.24\linewidth]{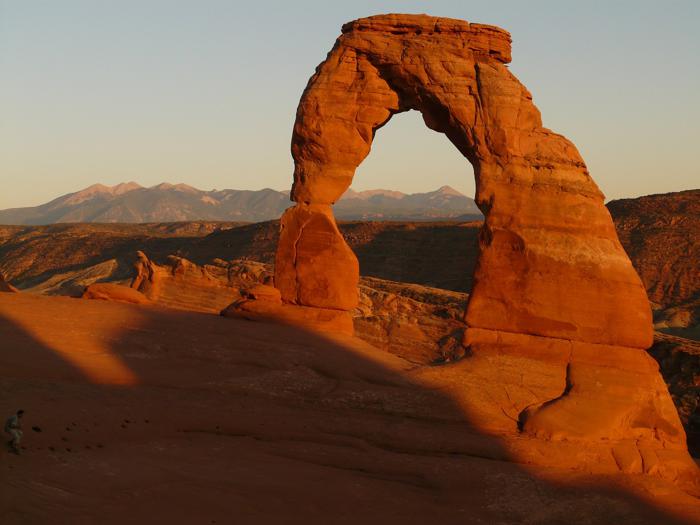}}}
\end{overpic}
\end{subfigure}
\begin{subfigure}[b]{0.185\linewidth}
\centering
Mask
\begin{overpic}[width=\linewidth, height=0.62\linewidth]{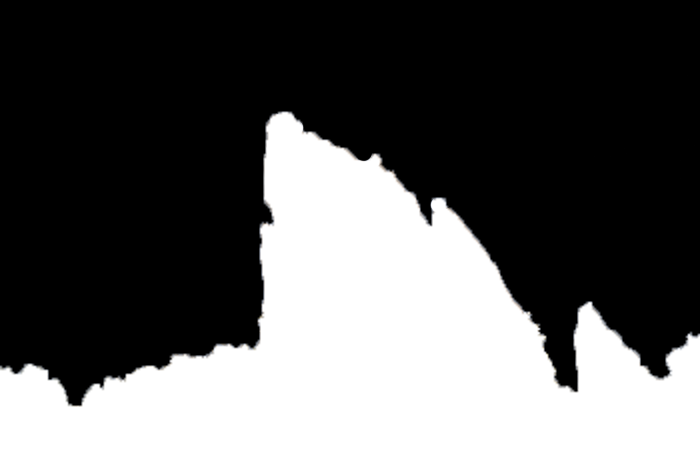}
\put(-2,38){\color{black}%
	\frame{\includegraphics[width=0.4\linewidth, height=0.24\linewidth]{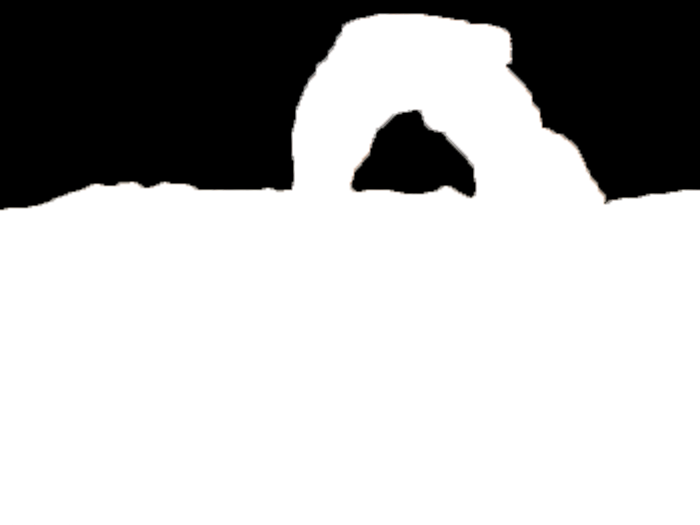}}}
\end{overpic}
\end{subfigure}
\begin{subfigure}[b]{0.185\linewidth}
\centering
Luan \cite{luan17}
\includegraphics[width=\linewidth, height=0.62\linewidth]{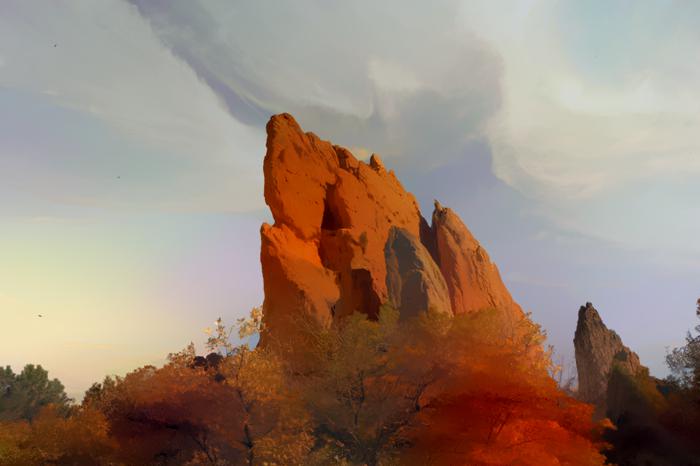}
\end{subfigure}
\begin{subfigure}[b]{0.185\linewidth}
\centering
Li \cite{li2018}
\includegraphics[width=\linewidth, height=0.62\linewidth]{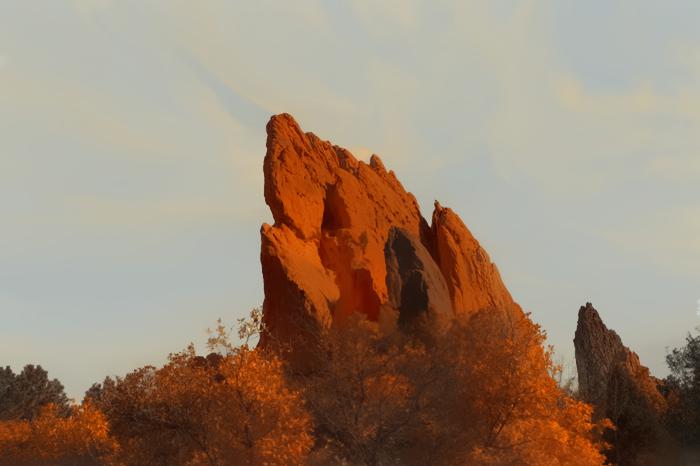}
\end{subfigure}
\begin{subfigure}[b]{0.185\linewidth}
\centering
Ours
\includegraphics[width=\linewidth, height=0.62\linewidth]{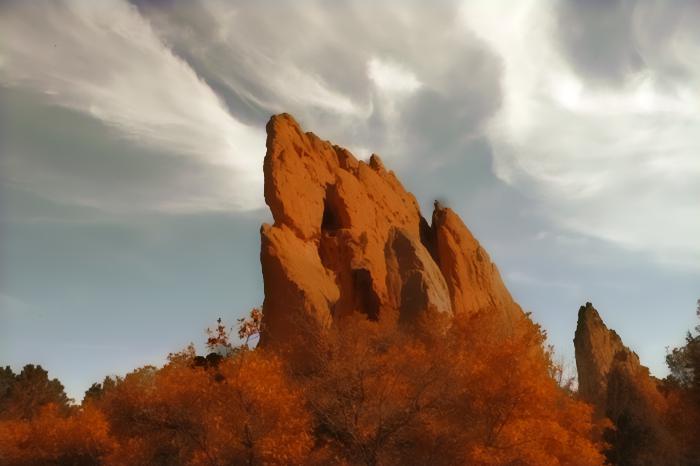}
\end{subfigure}
\vspace{1pt}

\hspace{5pt}
\begin{subfigure}[b]{0.185\linewidth}
\begin{overpic}[width=\linewidth, height=0.62\linewidth]{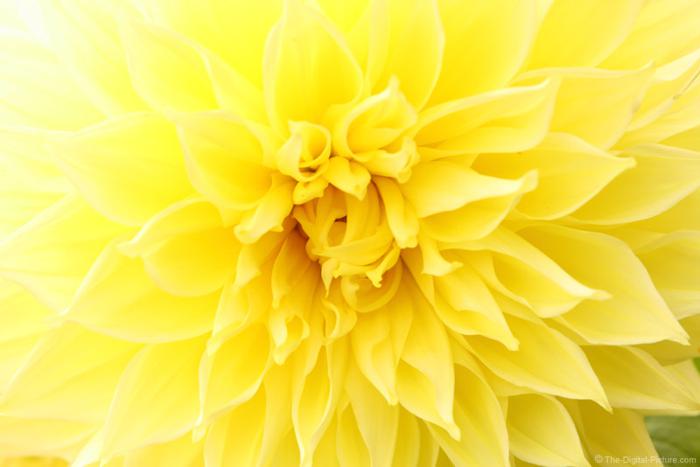}
\put(-2,38){\color{white}%
	\frame{\includegraphics[width=0.4\linewidth, height=0.24\linewidth]{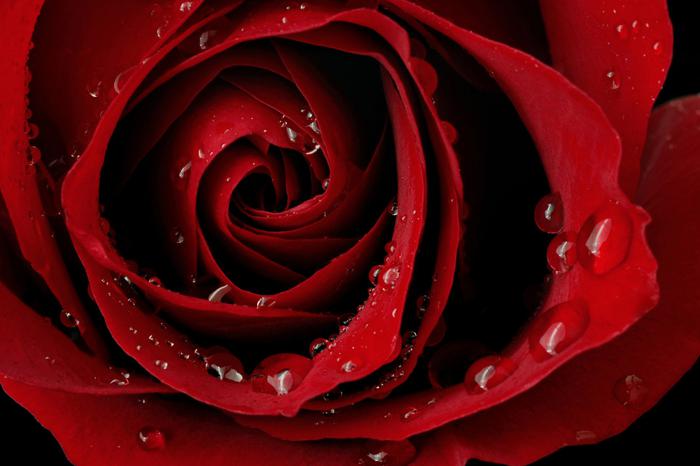}}}
\end{overpic}
\end{subfigure}
\begin{subfigure}[b]{0.185\linewidth}
\centering
\begin{overpic}[width=\linewidth, height=0.62\linewidth]{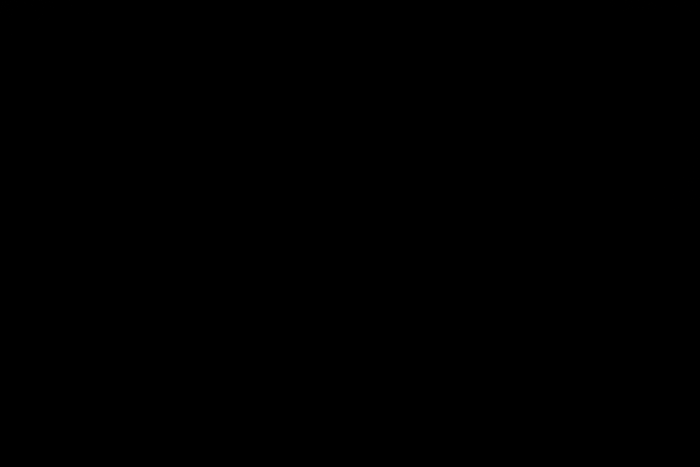}
\put(-2,38){\color{white}%
	\frame{\includegraphics[width=0.4\linewidth, height=0.24\linewidth]{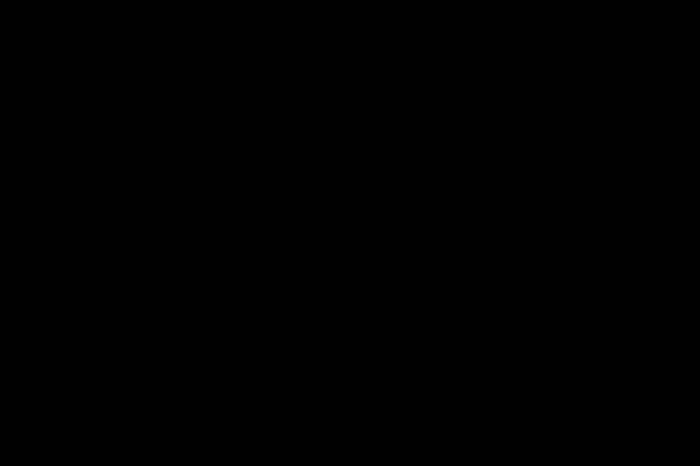}}}
\end{overpic}
\end{subfigure}
\begin{subfigure}[b]{0.185\linewidth}
\includegraphics[width=\linewidth, height=0.62\linewidth]{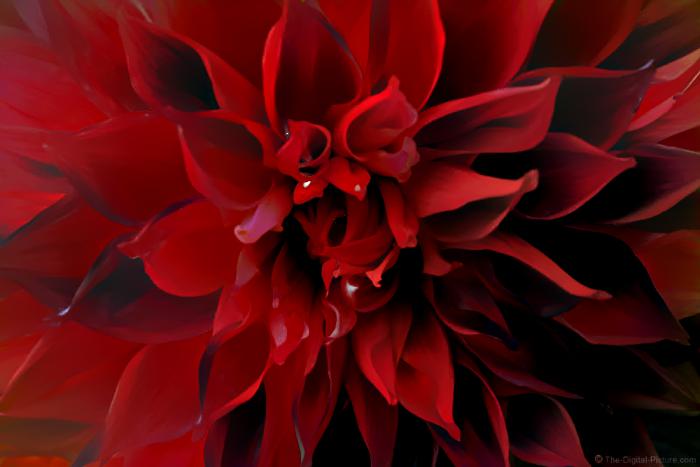}
\end{subfigure}
\begin{subfigure}[b]{0.185\linewidth}
\includegraphics[width=\linewidth, height=0.62\linewidth]{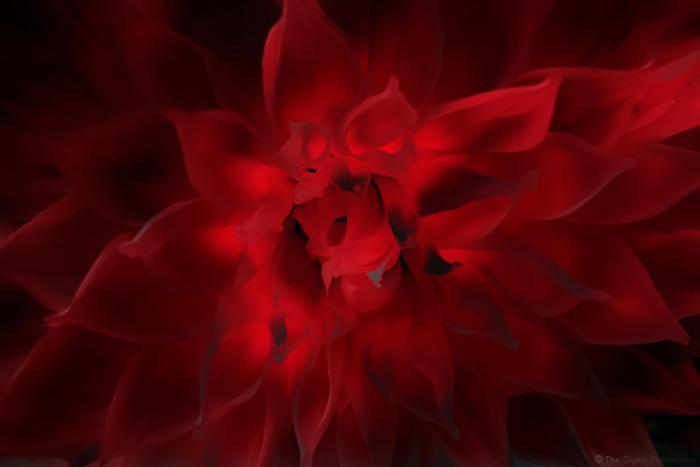}
\end{subfigure}
\begin{subfigure}[b]{0.185\linewidth}
\includegraphics[width=\linewidth, height=0.62\linewidth]{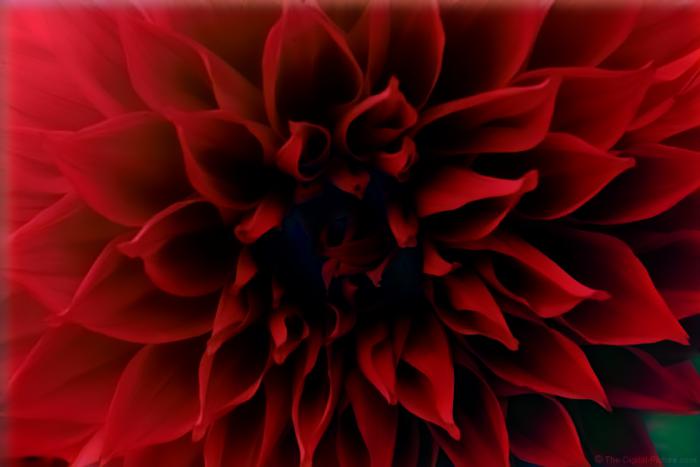}
\end{subfigure}
\vspace{1pt}

\hspace{5pt}
\begin{subfigure}[b]{0.185\linewidth}
\begin{overpic}[width=\linewidth, height=0.62\linewidth]{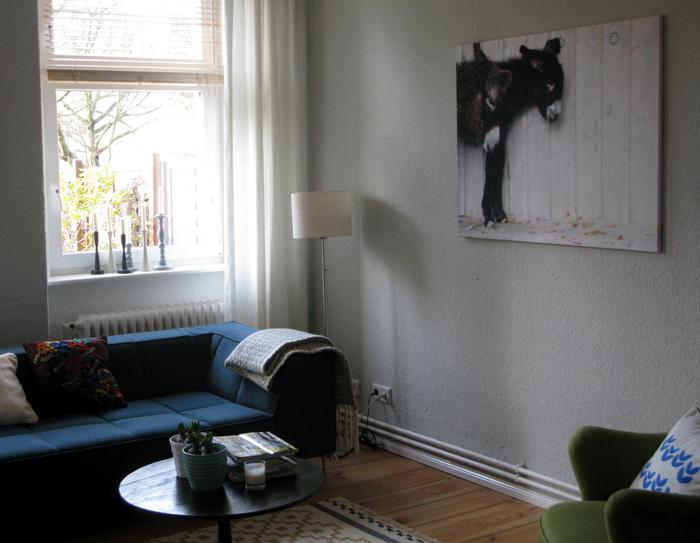}
\put(-2,38){\color{white}%
	\frame{\includegraphics[width=0.4\linewidth, height=0.24\linewidth]{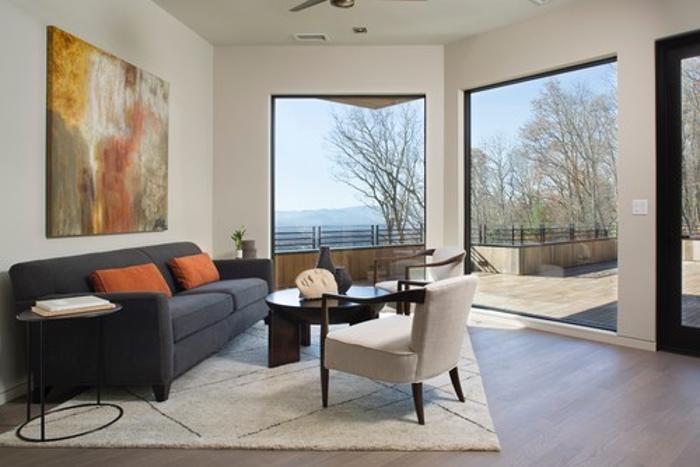}}}
\end{overpic}
\end{subfigure}
\begin{subfigure}[b]{0.185\linewidth}
\centering
\begin{overpic}[width=\linewidth, height=0.62\linewidth]{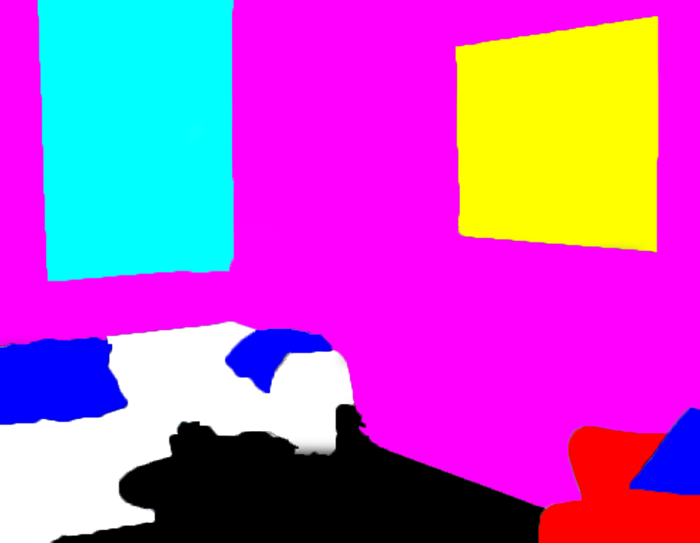}
\put(-2,38){\color{black}%
	\frame{\includegraphics[width=0.4\linewidth, height=0.24\linewidth]{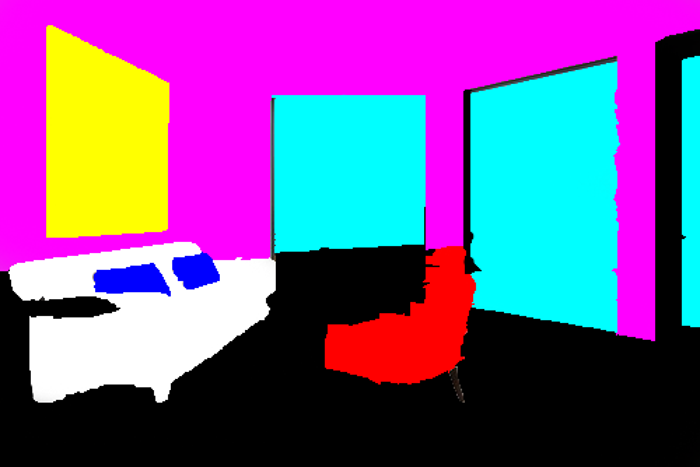}}}
\end{overpic}
\end{subfigure}
\begin{subfigure}[b]{0.185\linewidth}
\includegraphics[width=\linewidth, height=0.62\linewidth]{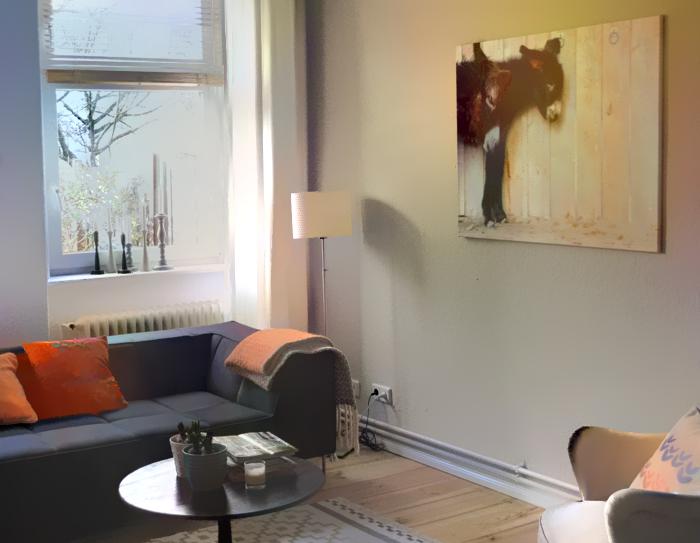}
\end{subfigure}
\begin{subfigure}[b]{0.185\linewidth}
\includegraphics[width=\linewidth, height=0.62\linewidth]{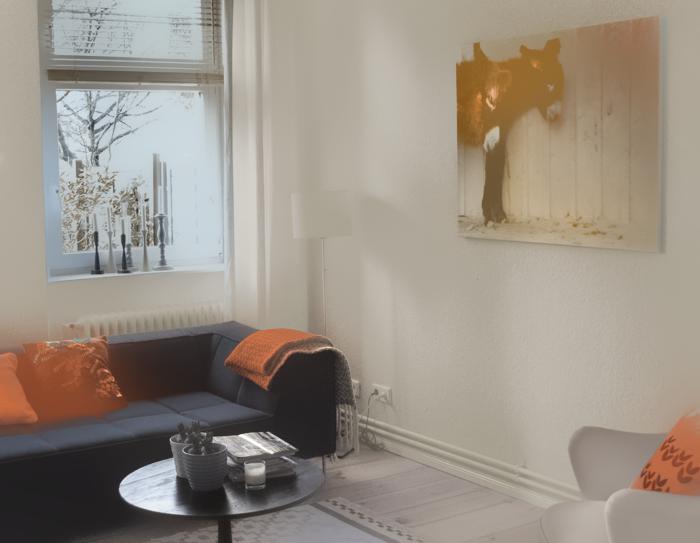}
\end{subfigure}
\begin{subfigure}[b]{0.185\linewidth}
\includegraphics[width=\linewidth, height=0.62\linewidth]{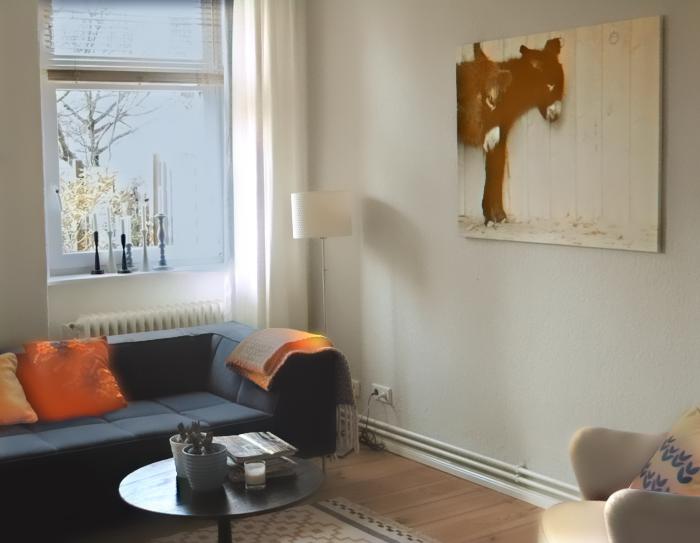}
\end{subfigure}
\vspace{1pt}

\hspace{5pt}
\begin{subfigure}[b]{0.185\linewidth}
\begin{overpic}[width=\linewidth, height=0.62\linewidth]{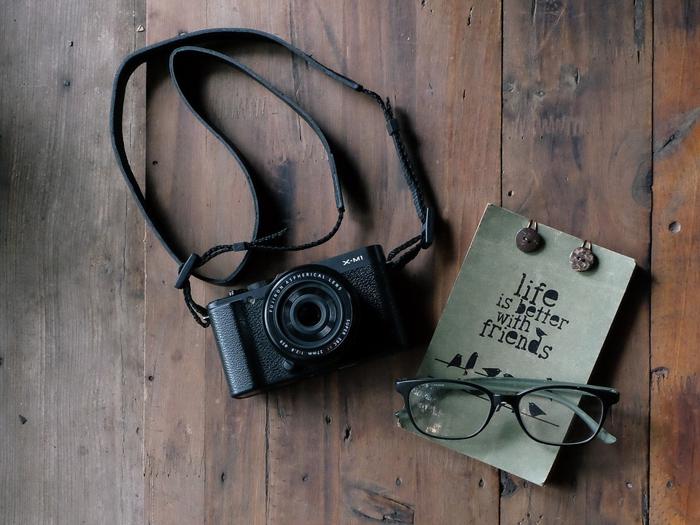}
\put(-2,38){\color{white}%
	\frame{\includegraphics[width=0.4\linewidth, height=0.24\linewidth]{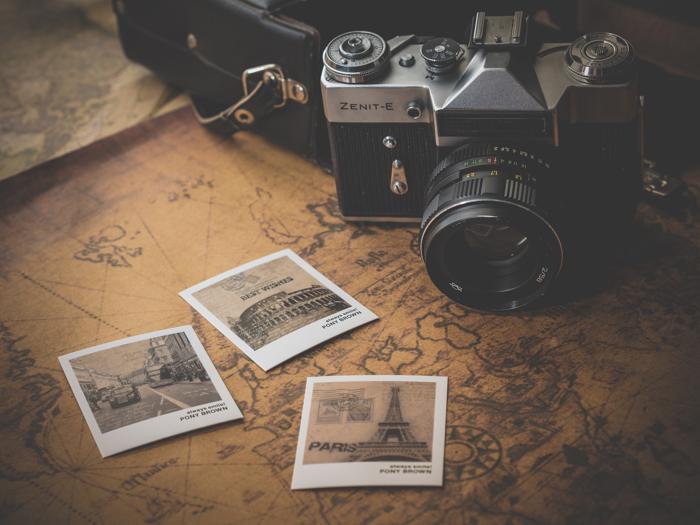}}}
\end{overpic}
\end{subfigure}
\begin{subfigure}[b]{0.185\linewidth}
\centering
\begin{overpic}[width=\linewidth, height=0.62\linewidth]{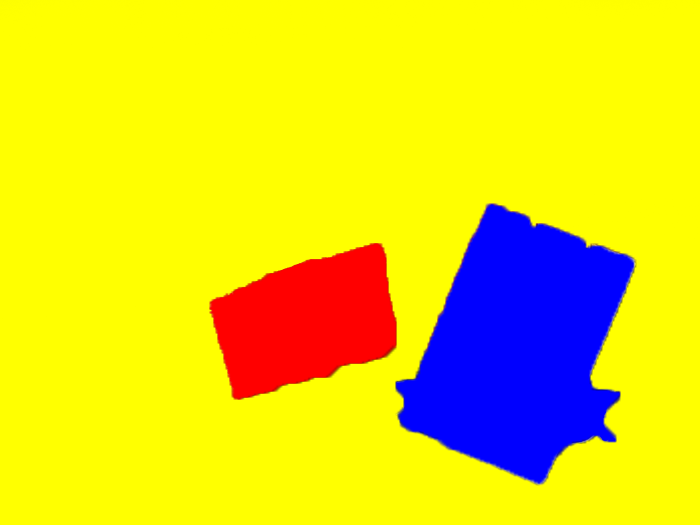}
\put(-2,38){\color{black}%
	\frame{\includegraphics[width=0.4\linewidth, height=0.24\linewidth]{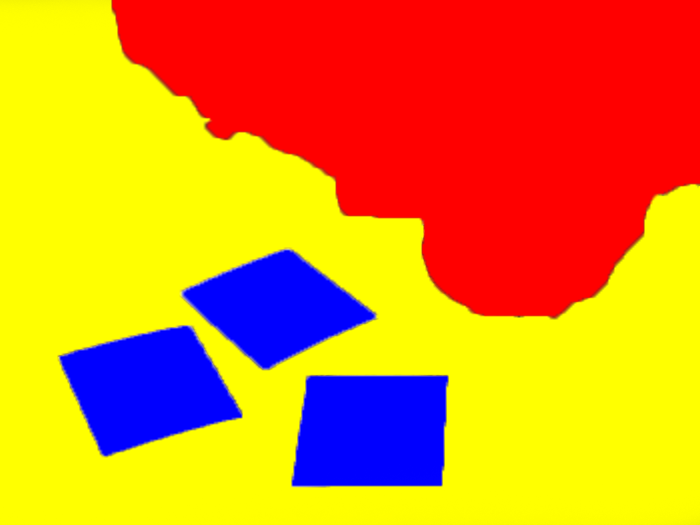}}}
\end{overpic}
\end{subfigure}
\begin{subfigure}[b]{0.185\linewidth}
\includegraphics[width=\linewidth, height=0.62\linewidth]{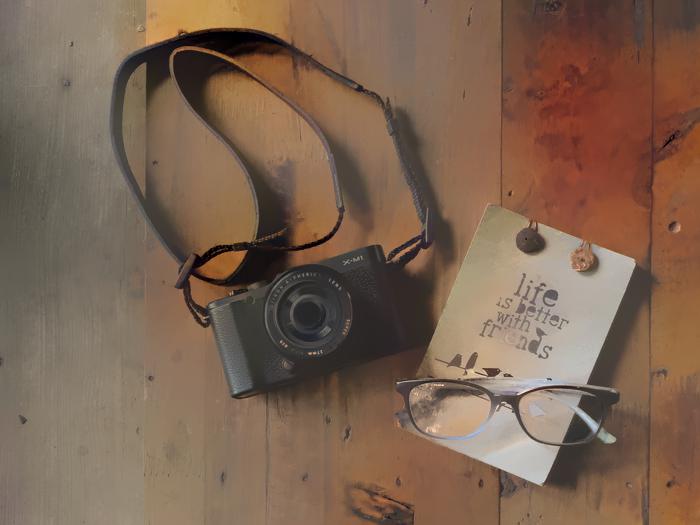}
\end{subfigure}
\begin{subfigure}[b]{0.185\linewidth}
\includegraphics[width=\linewidth, height=0.62\linewidth]{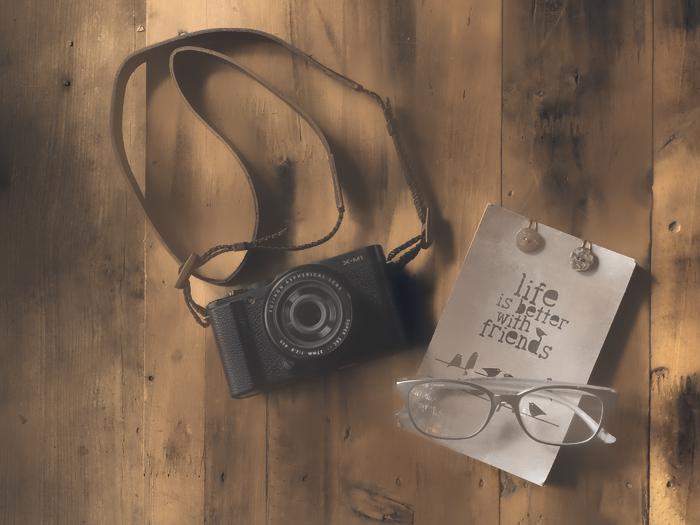}
\end{subfigure}
\begin{subfigure}[b]{0.185\linewidth}
\includegraphics[width=\linewidth, height=0.62\linewidth]{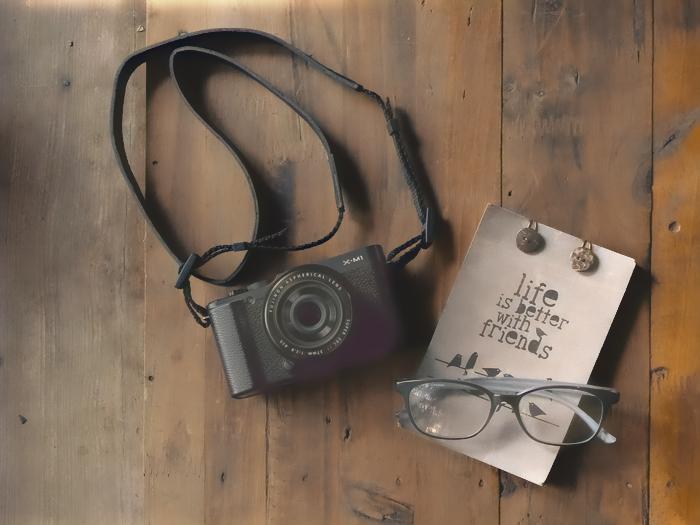}
\end{subfigure}
\vspace{1pt}

\hspace{5pt}
\begin{subfigure}[b]{0.185\linewidth}
\begin{overpic}[width=\linewidth, height=0.62\linewidth]{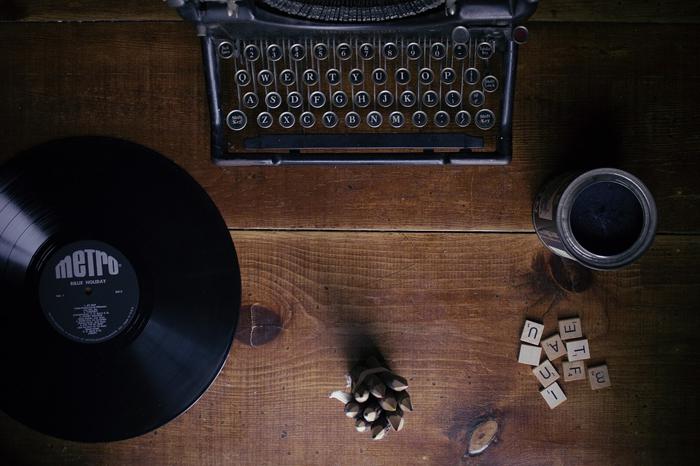}
\put(-2,38){\color{white}%
	\frame{\includegraphics[width=0.4\linewidth, height=0.24\linewidth]{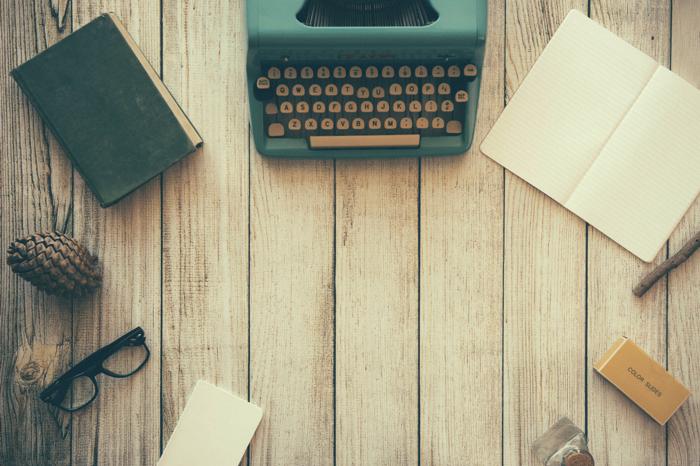}}}
\end{overpic}
\end{subfigure}
\begin{subfigure}[b]{0.185\linewidth}
\centering
\begin{overpic}[width=\linewidth, height=0.62\linewidth]{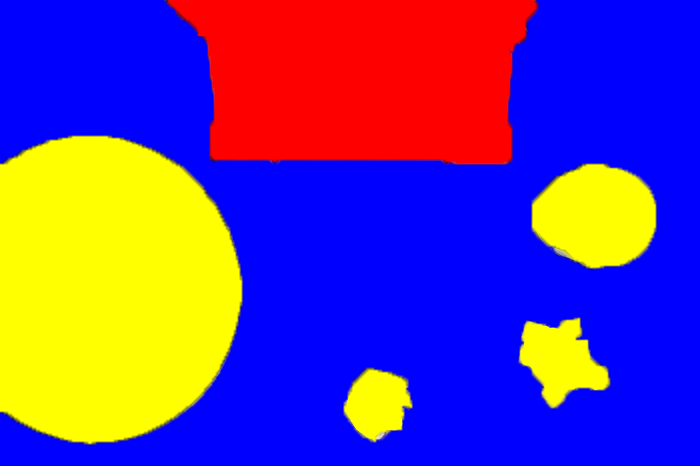}
\put(-2,38){\color{black}%
	\frame{\includegraphics[width=0.4\linewidth, height=0.24\linewidth]{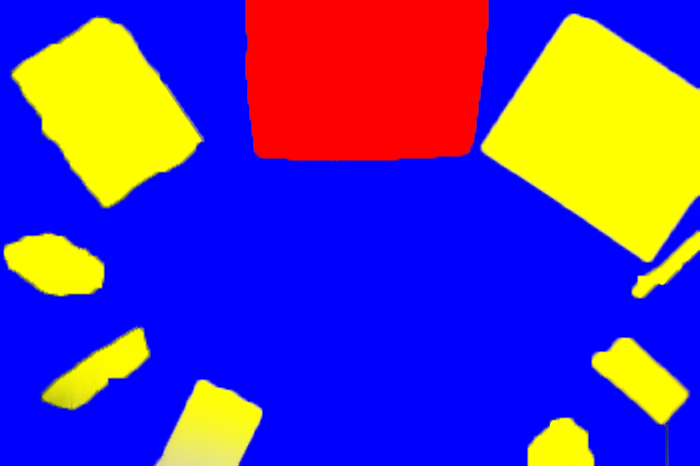}}}
\end{overpic}
\end{subfigure}
\begin{subfigure}[b]{0.185\linewidth}
\includegraphics[width=\linewidth, height=0.62\linewidth]{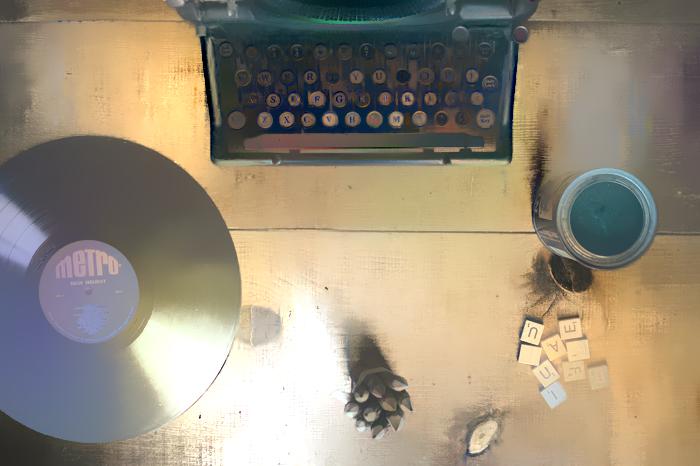}
\end{subfigure}
\begin{subfigure}[b]{0.185\linewidth}
\includegraphics[width=\linewidth, height=0.62\linewidth]{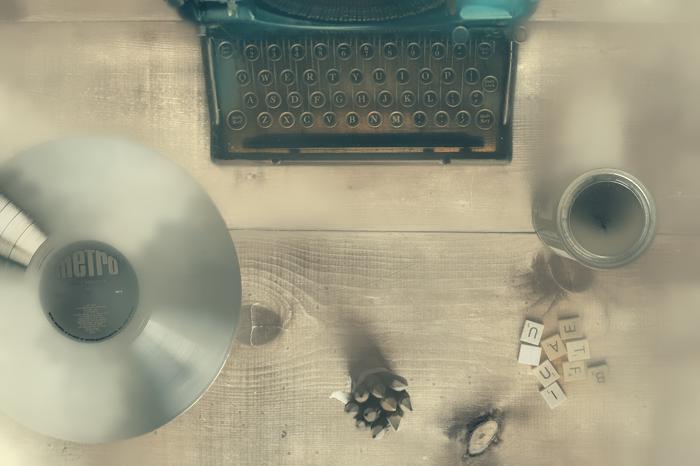}
\end{subfigure}
\begin{subfigure}[b]{0.185\linewidth}
\includegraphics[width=\linewidth, height=0.62\linewidth]{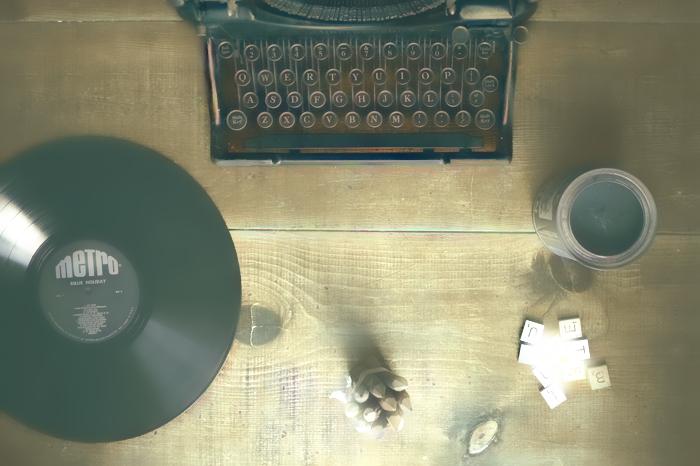}
\end{subfigure}
\vspace{1pt}

\hspace{5pt}
\begin{subfigure}[b]{0.185\linewidth}
\begin{overpic}[width=\linewidth, height=0.62\linewidth]{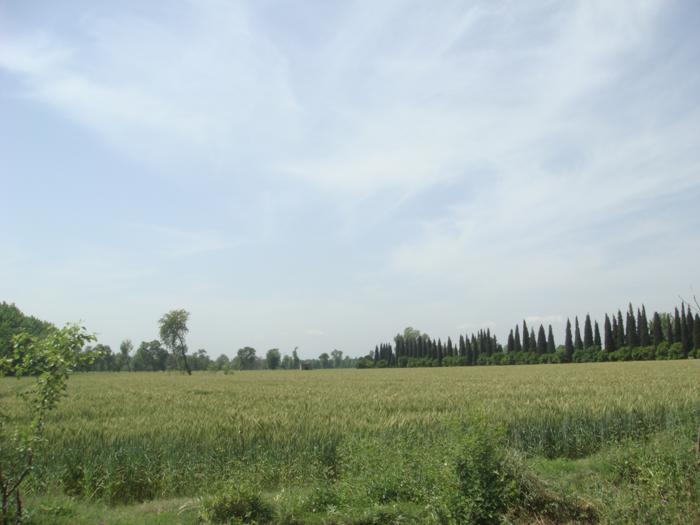}
\put(-2,38){\color{white}%
	\frame{\includegraphics[width=0.4\linewidth, height=0.24\linewidth]{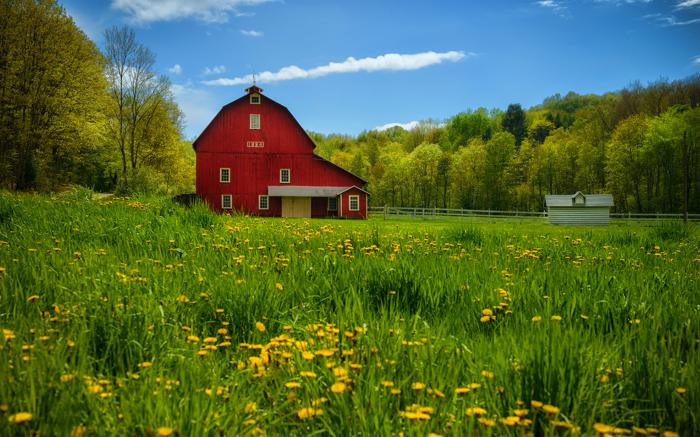}}}
\end{overpic}
\end{subfigure}
\begin{subfigure}[b]{0.185\linewidth}
\centering
\begin{overpic}[width=\linewidth, height=0.62\linewidth]{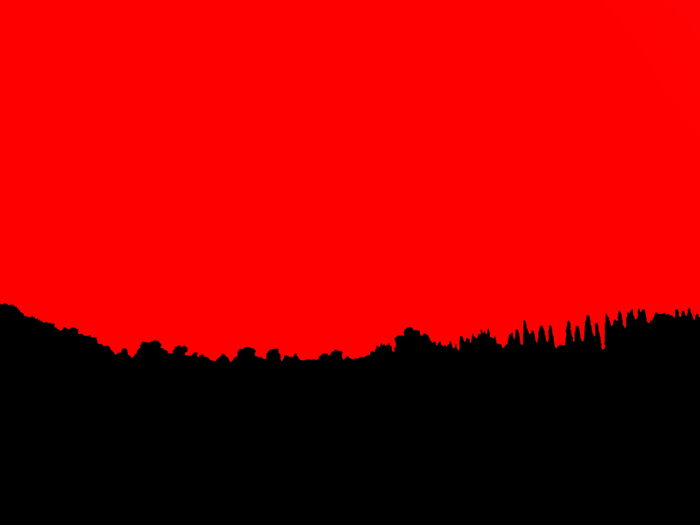}
\put(-2,38){\color{black}%
	\frame{\includegraphics[width=0.4\linewidth, height=0.24\linewidth]{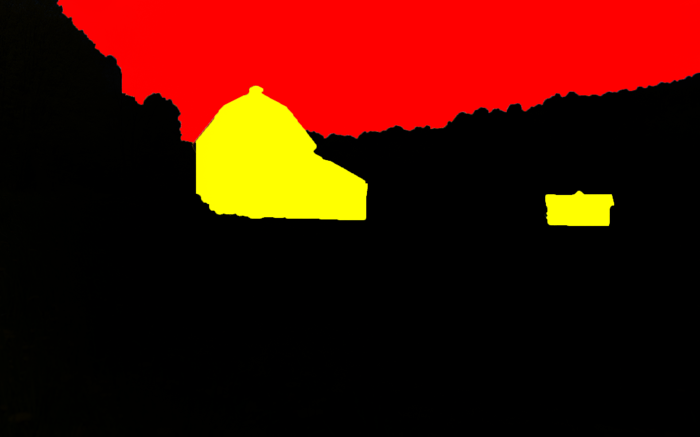}}}
\end{overpic}
\end{subfigure}
\begin{subfigure}[b]{0.185\linewidth}
\includegraphics[width=\linewidth, height=0.62\linewidth]{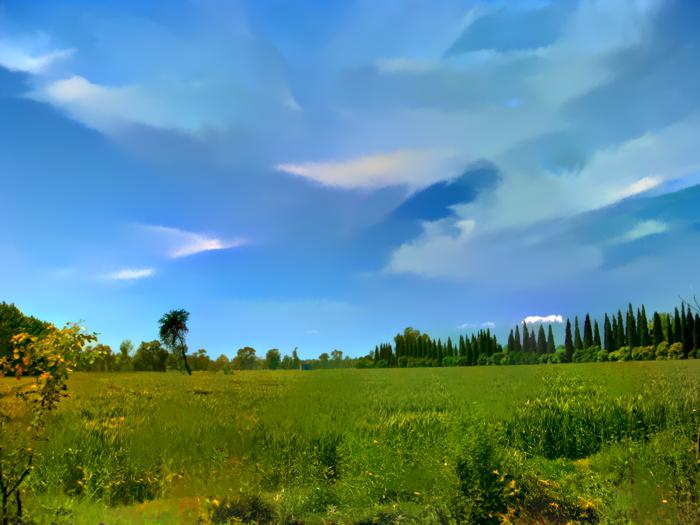}
\end{subfigure}
\begin{subfigure}[b]{0.185\linewidth}
\includegraphics[width=\linewidth, height=0.62\linewidth]{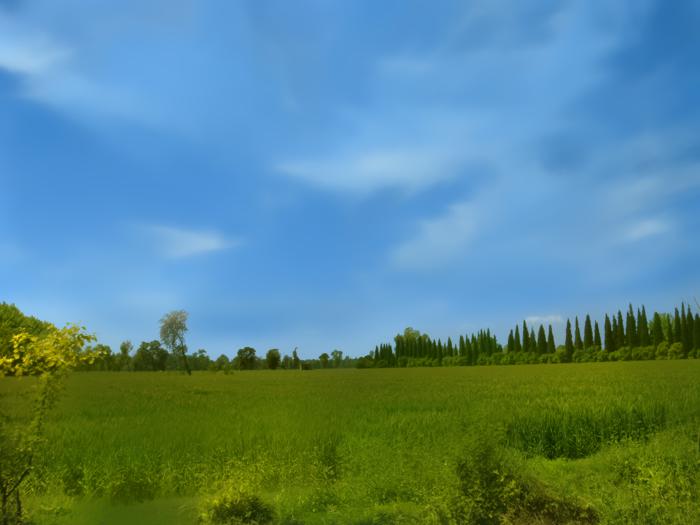}
\end{subfigure}
\begin{subfigure}[b]{0.185\linewidth}
\includegraphics[width=\linewidth, height=0.62\linewidth]{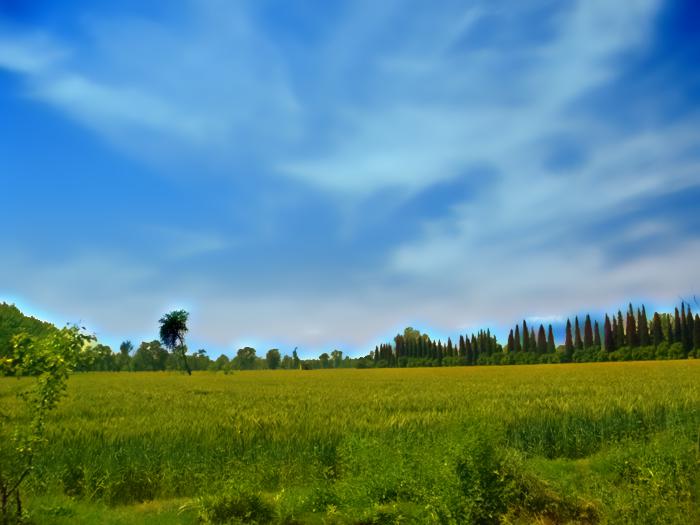}
\end{subfigure}
\vspace{3pt}

\hrule\ 
\vspace{2pt}

\rotatebox{90}{\scriptsize{Failure case}}
\begin{subfigure}[b]{0.185\linewidth}
\begin{overpic}[width=\linewidth, height=0.62\linewidth]{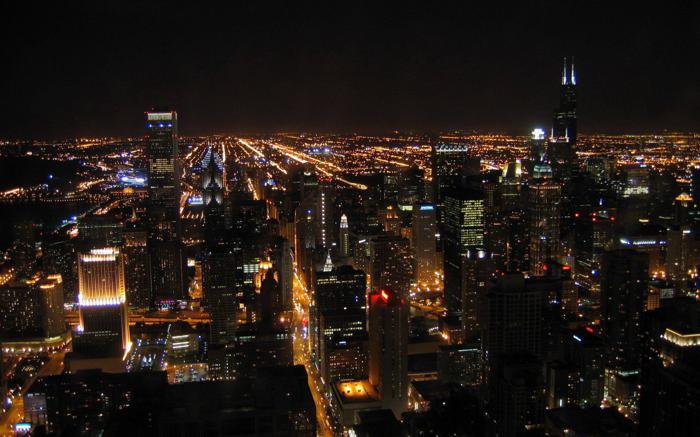}
\put(-2,38){\color{white}%
	\frame{\includegraphics[width=0.4\linewidth, height=0.24\linewidth]{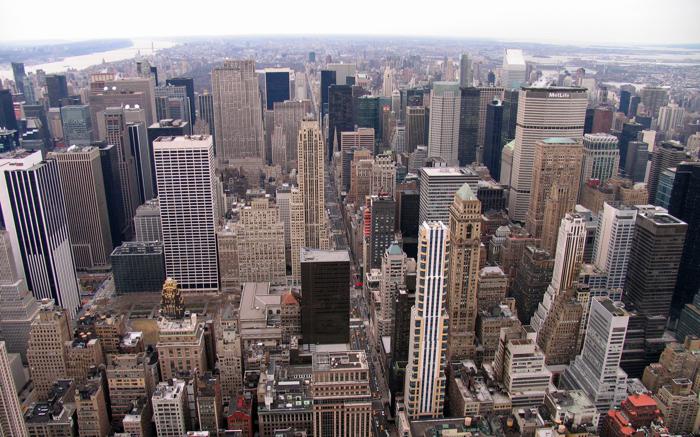}}}
\end{overpic}
\end{subfigure}
\begin{subfigure}[b]{0.185\linewidth}
\centering
\begin{overpic}[width=\linewidth, height=0.62\linewidth]{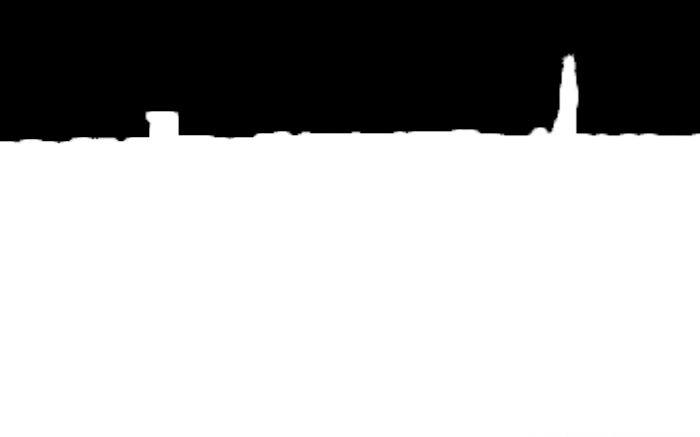}
\put(-2,38){\color{black}%
	\frame{\includegraphics[width=0.4\linewidth, height=0.24\linewidth]{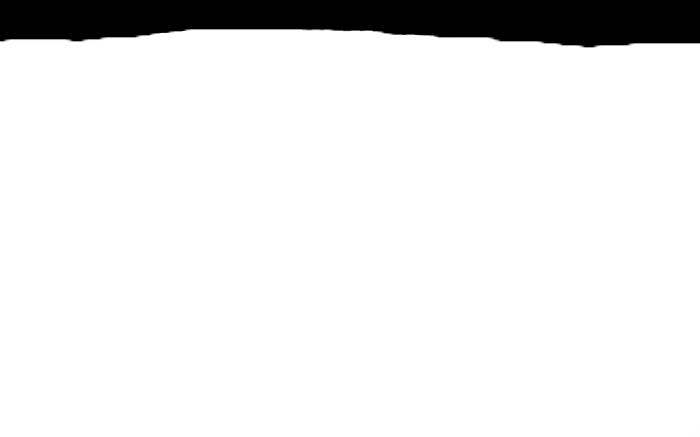}}}
\end{overpic}
\end{subfigure}
\begin{subfigure}[b]{0.185\linewidth}
\includegraphics[width=\linewidth, height=0.62\linewidth]{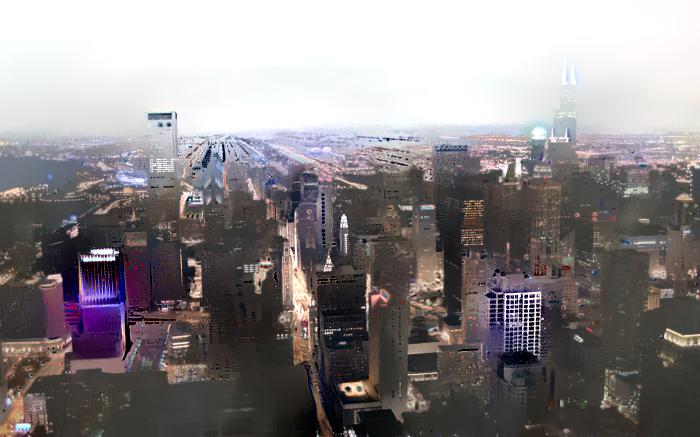}
\end{subfigure}
\begin{subfigure}[b]{0.185\linewidth}
\includegraphics[width=\linewidth, height=0.62\linewidth]{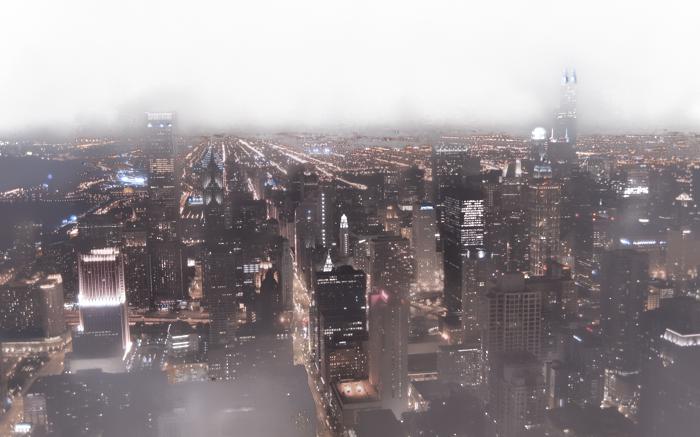}
\end{subfigure}
\begin{subfigure}[b]{0.185\linewidth}
\includegraphics[width=\linewidth, height=0.62\linewidth]{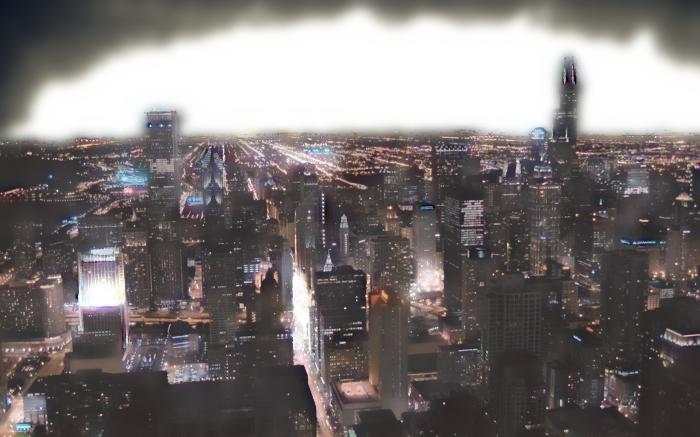}
\end{subfigure}
\vspace{1pt}

\end{center}
   \caption{Photo style transfer with the methods of \cite{luan17}, \cite{li2018} and ours.}
\label{compares}
\end{figure}

We compare our results with those of \cite{luan17, li2018} in Fig.~\ref{compares}. The method of \cite{luan17} has been described above. Li~\etal\cite{li2018} proposed a closed-form solution to the photo style transfer problem by using a feed-forward network for stylization followed by a smoothing step that favors photorealism. 

Qualitatively, the results of \cite{luan17} seem satisfying at first glance, but some of them present watercolor painting-like artifacts. The method \cite{li2018} is fast and preserves well the content structure. Nevertheless, the smoothing step tends to weaken the stylization, with a lack of color saturation in the output image compared to the style image, and sometimes yields an undesired haze effect.
In comparison, we find that our results respect better the original style while looking closer to real photos. They do not suffer from watercolor painting-like or hazy artifacts. However, we remarked that our approach sometimes generates inconsistent stylizations at the boundaries of different semantic regions. We also noticed that it has difficulties transferring city landscape images from night to day. We show such a failure case in the last row of Fig.~\ref{compares}. Note that the results obtained by \cite{luan17} or \cite{li2018} are also not entirely satisfying for this example. Finally, we remarked that the method of \cite{li2018} has more difficulties to transfer the color saturation of the style image than the method of \cite{luan17} or ours. A measure of the distance between the histograms (computed in the channel S of the HSV color model) of the style images and stylized images shows that, on average, \cite{luan17} preserves the best color saturation followed by our method.

\subsection{Generalization to unseen images}
\begin{figure}
\begin{center}
\begin{subfigure}[b]{0.24\linewidth}
\begin{overpic}[width=\linewidth, height=0.62\linewidth]{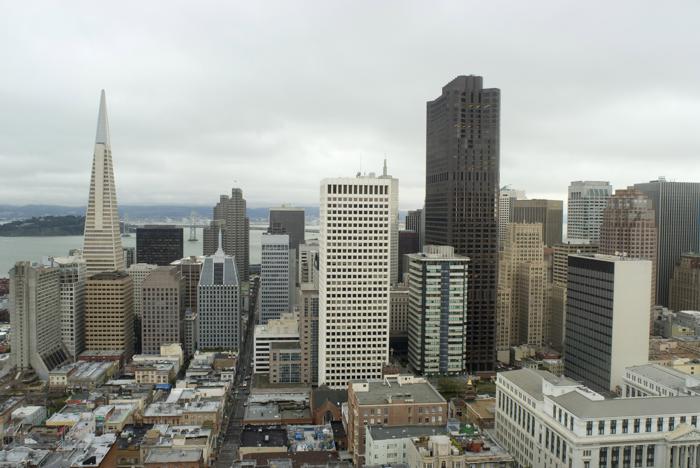}
\put(-2,38){\color{white}%
	\frame{\includegraphics[width=0.4\linewidth, height=0.25\linewidth]{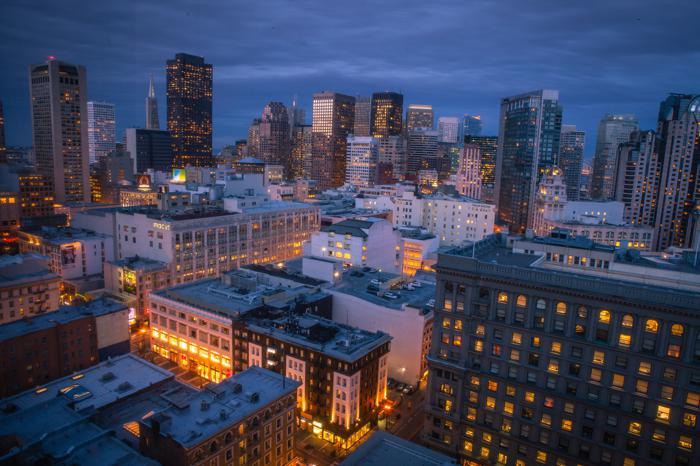}}}
\end{overpic}

\vspace{0.05cm}
\begin{overpic}[width=\linewidth, height=0.62\linewidth]{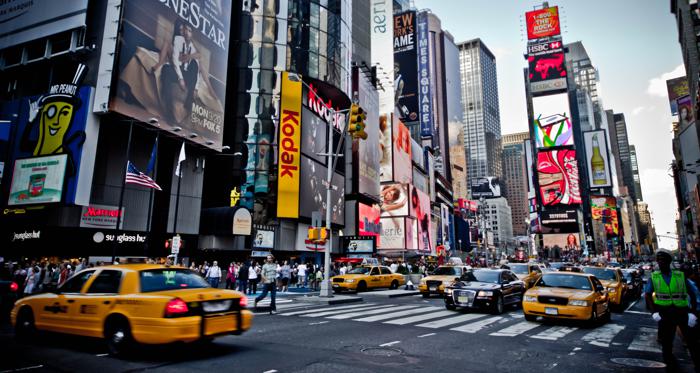}
\put(-2,38){\color{white}%
	\frame{\includegraphics[width=0.4\linewidth, height=0.25\linewidth]{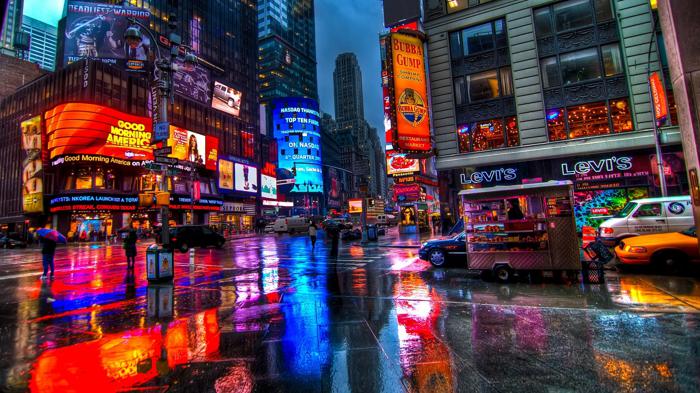}}}
\end{overpic}

\vspace{0.05cm}
\begin{overpic}[width=\linewidth, height=0.62\linewidth]{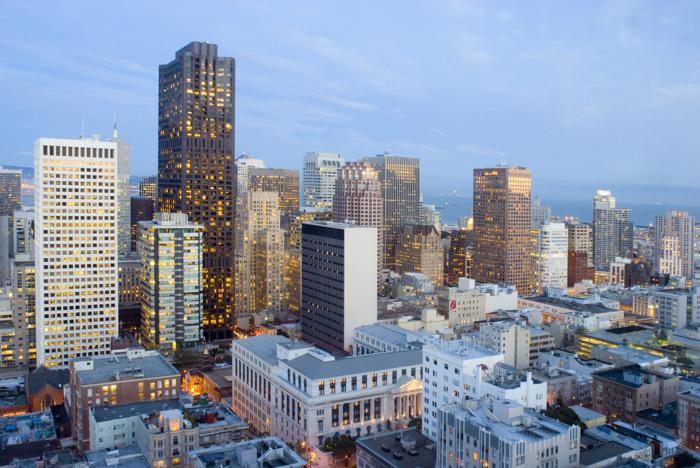}
\put(-2,38){\color{white}%
	\frame{\includegraphics[width=0.4\linewidth, height=0.25\linewidth]{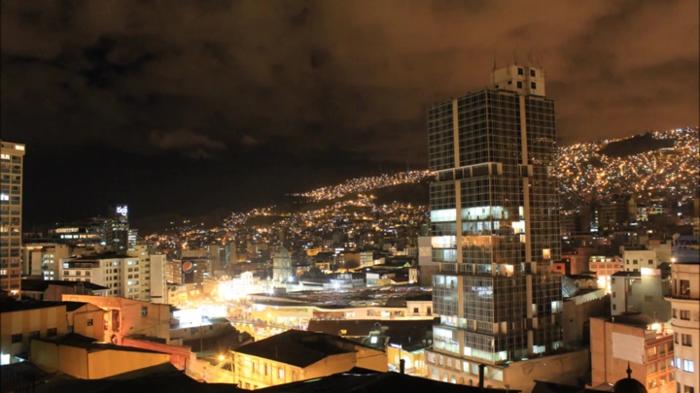}}}
\end{overpic}
\caption{\small Pre-training}
\end{subfigure}
\begin{subfigure}[b]{0.24\linewidth}
\includegraphics[width=\linewidth, height=0.62\linewidth]{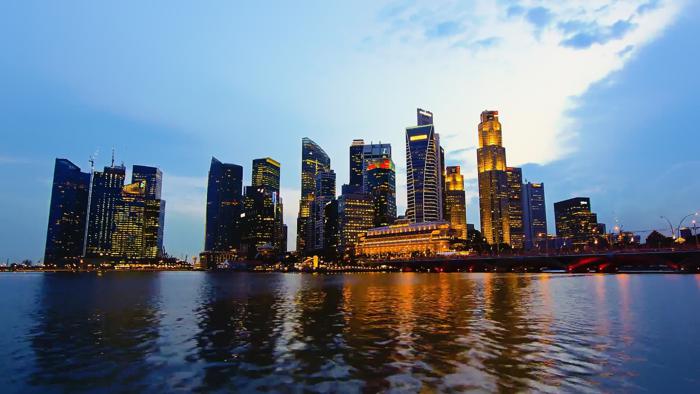}

\vspace{0.05cm}
\includegraphics[width=\linewidth, height=0.62\linewidth]{jpg/content/in19.jpg}

\vspace{0.05cm}
\includegraphics[width=\linewidth, height=0.62\linewidth]{jpg/content/in10.jpg}
\caption{\small Input}
\end{subfigure}
\begin{subfigure}[b]{0.24\linewidth}
\includegraphics[width=\linewidth, height=0.62\linewidth]{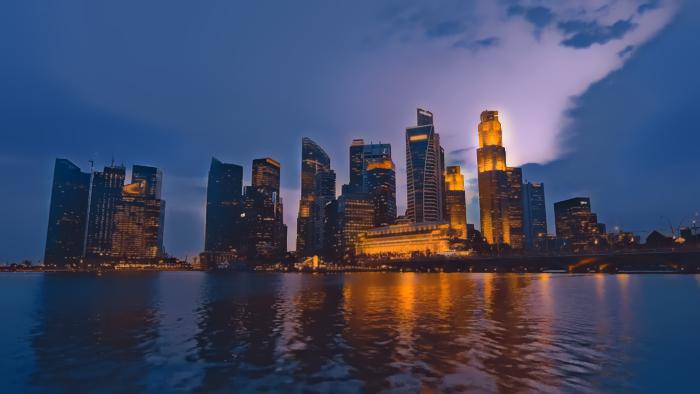}

\vspace{0.05cm}
\includegraphics[width=\linewidth, height=0.62\linewidth]{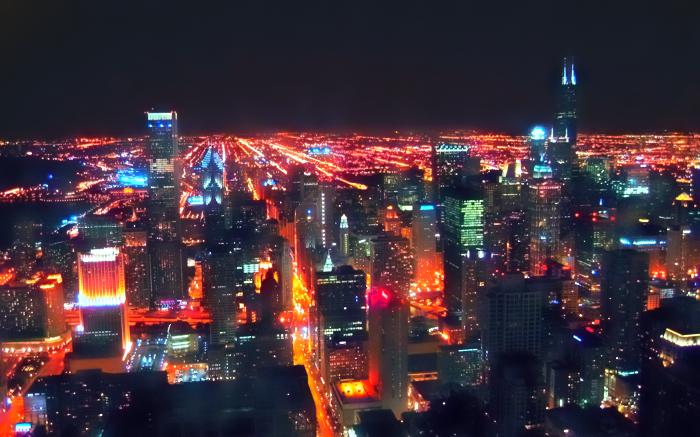}

\vspace{0.05cm}
\includegraphics[width=\linewidth, height=0.62\linewidth]{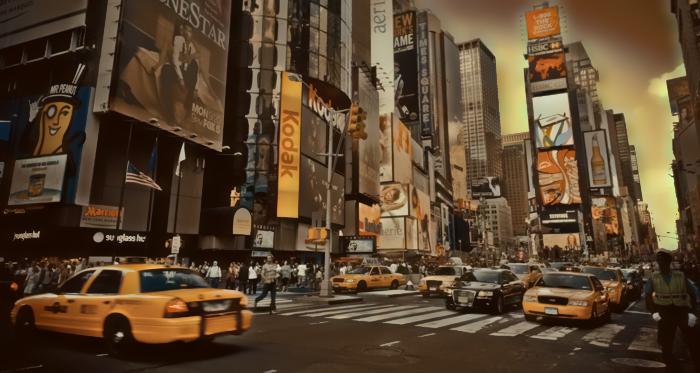}
\caption{\small Reapplied}
\end{subfigure}
\begin{subfigure}[b]{0.24\linewidth}
\includegraphics[width=\linewidth, height=0.62\linewidth]{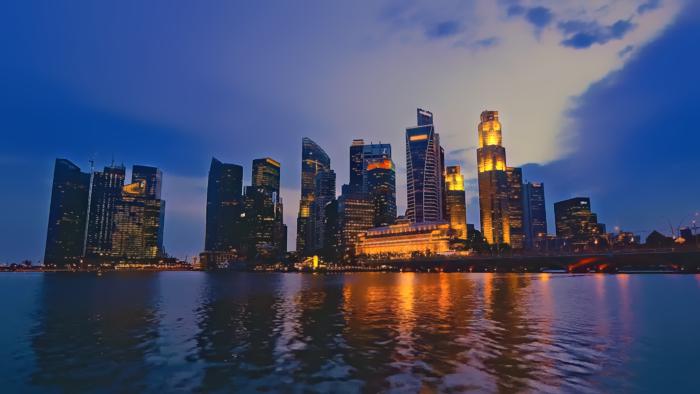}

\vspace{0.05cm}
\includegraphics[width=\linewidth, height=0.62\linewidth]{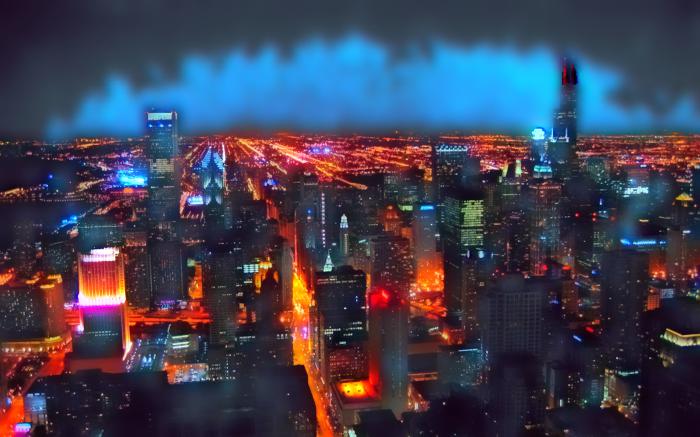}

\vspace{0.05cm}
\includegraphics[width=\linewidth, height=0.62\linewidth]{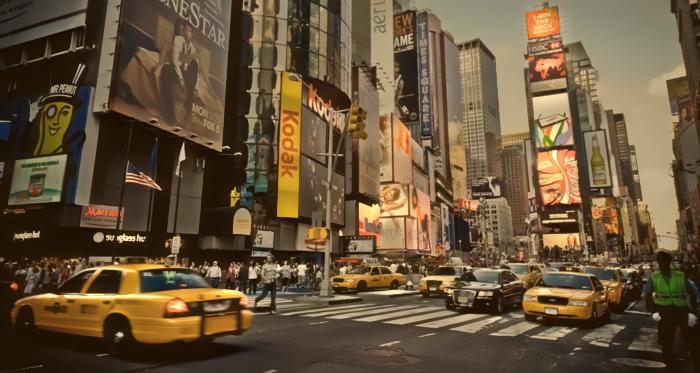}
\caption{\small Optimized}
\end{subfigure}
\end{center}
   \caption{Style transfer networks pre-trained on (a) have been reapplied on (b), getting stylized images (c). (d) are outputs generated by our basic approach using (b) and the small top image of (a) as reference image. }
\label{reapply}
\end{figure}

Even though trained on merely two images, our network can stylize images not viewed at training time. As an example, we first train style transfer networks on the pair of images in Fig.~\ref{reapply}a. We then use these trained networks to transfer the styles of the small top images in Fig.~\ref{reapply}a to the images in Fig.~\ref{reapply}b. We obtained the stylized images in Fig.~\ref{reapply}c, which preserve well the original structures while incorporating the target styles. 
For comparison, we show the results obtained with our original approach in Fig.~\ref{reapply}d, \emph{i.e.}, obtained after training new convnets for each image in Fig.~\ref{reapply}b.  The results are globally comparable to the results obtained using the pre-trained convnets. We nevertheless remarked that to get a reasonable stylization one should use images of similar semantic content as the one used to pre-train the style transfer networks. When there is a semantic difference, our approach still transfers the styles but may generate unrealistic results, such as blue house or red river. For the domelike artifact in Fig.~\ref{reapply}d, there exist nearly invisible color differences in the dark area, so the network fails to transform all the area into blue.

\subsection{Retraining for new styles}
\begin{figure}
\captionsetup[subfigure]{font=scriptsize}
\begin{center}

\hspace{0.39\linewidth}
\begin{subfigure}[t]{0.19\linewidth}
\begin{overpic}[width=\linewidth, height=0.62\linewidth]{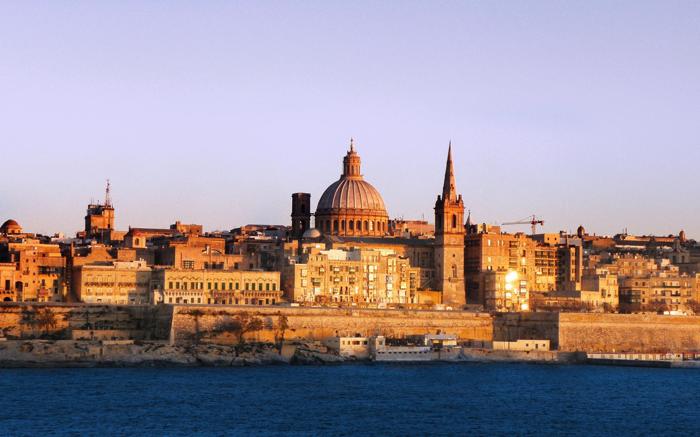}
\put(-2,38){\color{white}%
	\frame{\includegraphics[width=0.4\linewidth, height=0.25\linewidth]{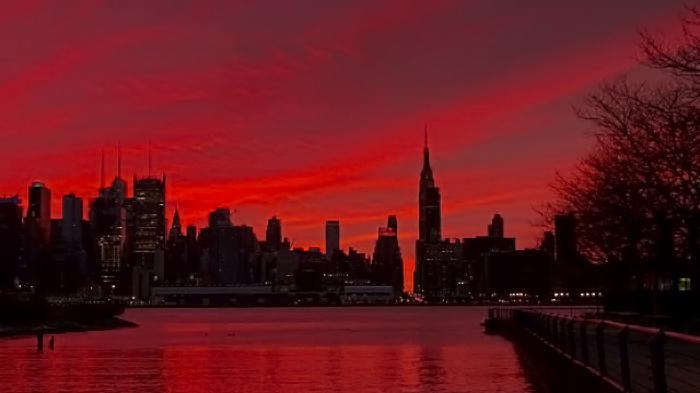}}}
\end{overpic}
\end{subfigure}
\begin{subfigure}[t]{0.19\linewidth}
\begin{overpic}[width=\linewidth, height=0.62\linewidth]{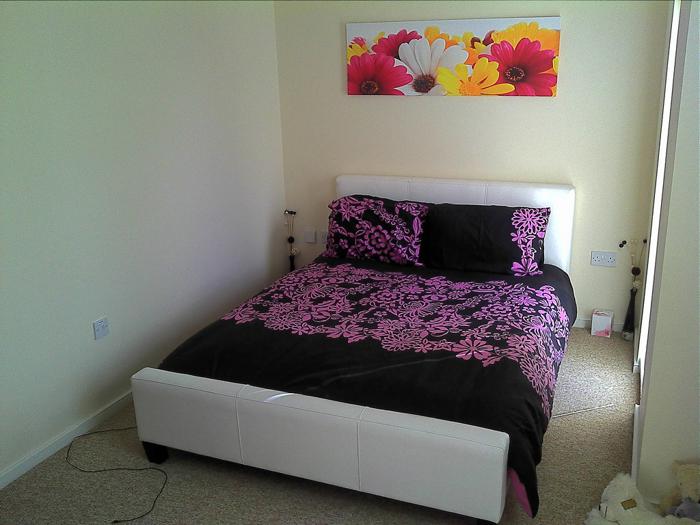}
\put(-2,38){\color{white}%
	\frame{\includegraphics[width=0.4\linewidth, height=0.25\linewidth]{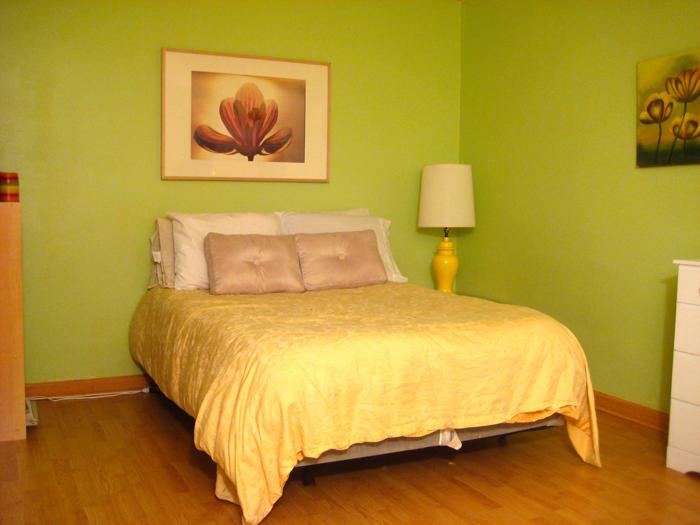}}}
\end{overpic}
\end{subfigure}
\begin{subfigure}[t]{0.19\linewidth}
\begin{overpic}[width=\linewidth, height=0.62\linewidth]{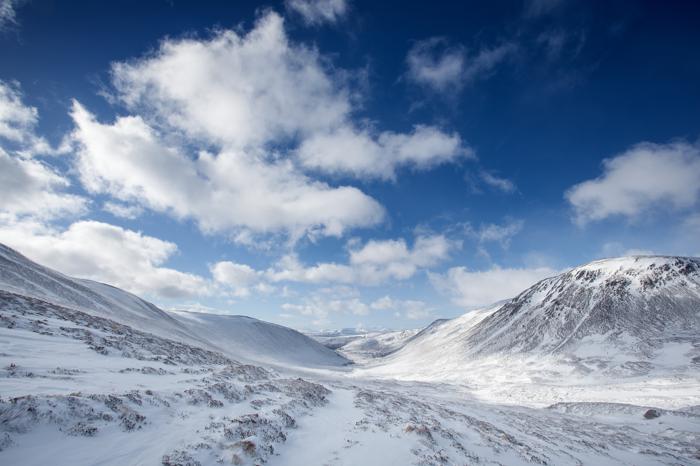}
\put(-2,38){\color{white}%
	\frame{\includegraphics[width=0.4\linewidth, height=0.25\linewidth]{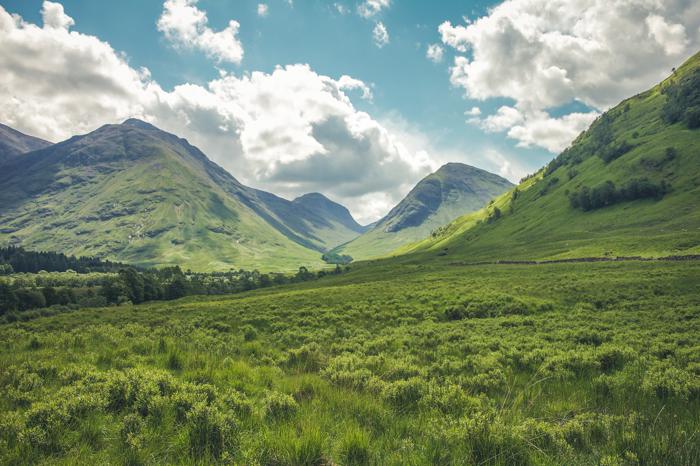}}}
\end{overpic}
\end{subfigure}
\vspace{1pt}

\begin{subfigure}[t]{0.19\linewidth}
\begin{overpic}[width=\linewidth, height=0.62\linewidth]{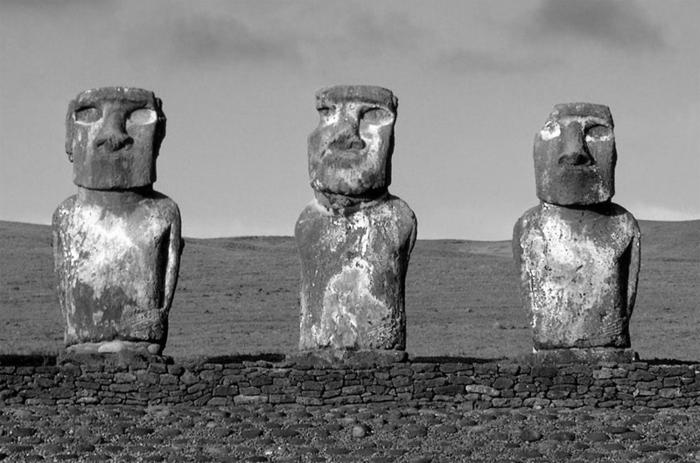}
\put(-2,38){\color{white}%
	\frame{\includegraphics[width=0.4\linewidth, height=0.25\linewidth]{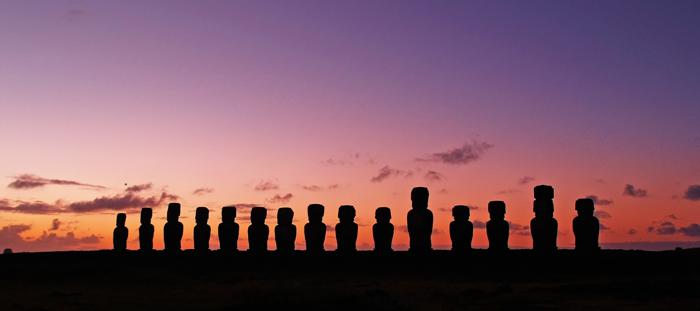}}}
\end{overpic}
\end{subfigure}
\begin{subfigure}[t]{0.19\linewidth}
\includegraphics[width=\linewidth, height=0.62\linewidth]{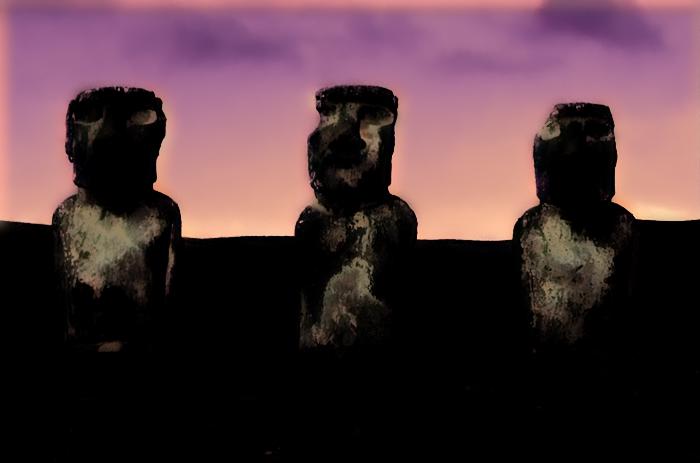}
\end{subfigure}
\begin{subfigure}[t]{0.19\linewidth}
\includegraphics[width=\linewidth, height=0.62\linewidth]{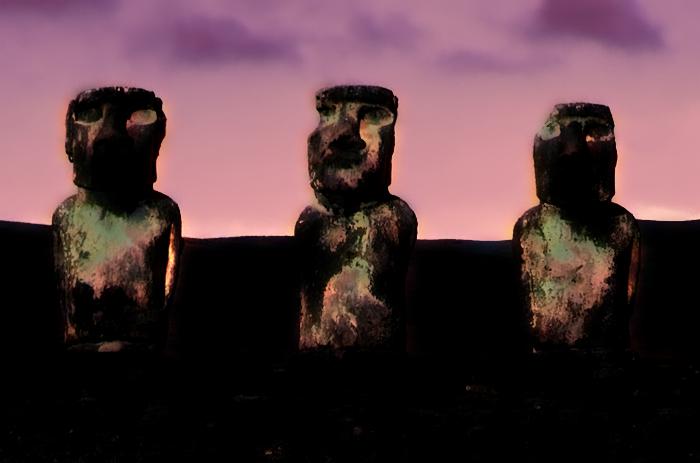}
\end{subfigure}
\begin{subfigure}[t]{0.19\linewidth}
\includegraphics[width=\linewidth, height=0.62\linewidth]{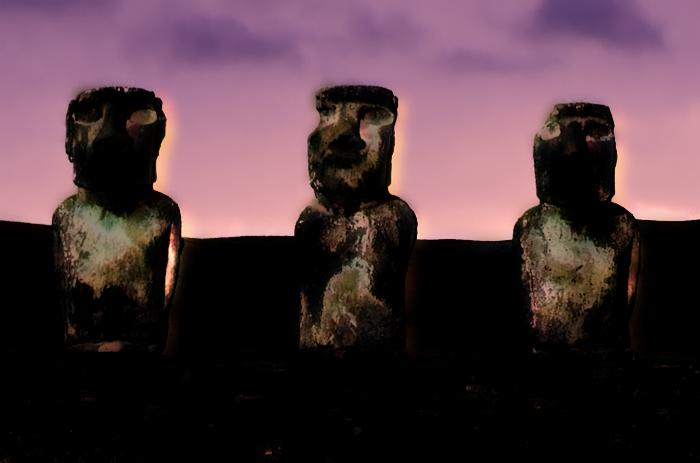}
\end{subfigure}
\begin{subfigure}[t]{0.19\linewidth}
\includegraphics[width=\linewidth, height=0.62\linewidth]{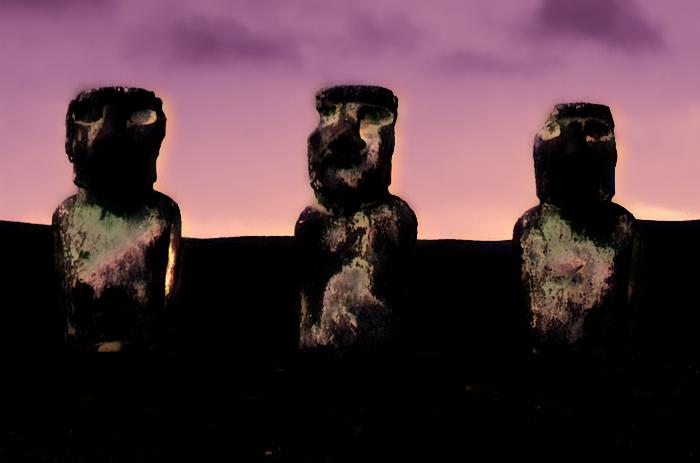}
\end{subfigure}
\vspace{1pt}

\begin{subfigure}[t]{0.19\linewidth}
\begin{overpic}[width=\linewidth, height=0.62\linewidth]{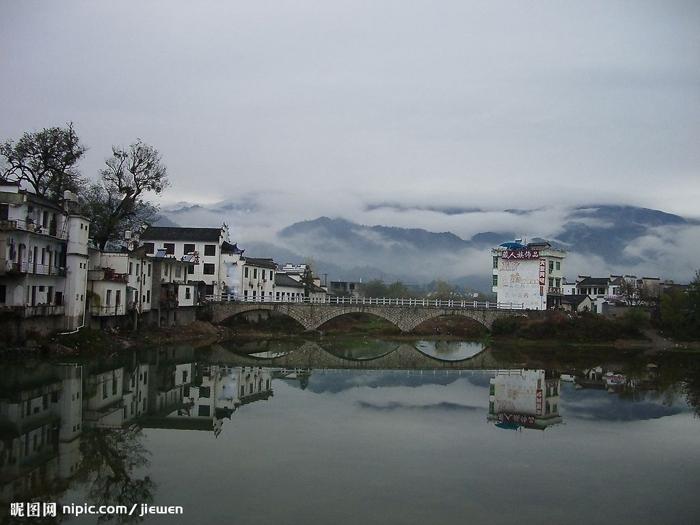}
\put(-2,38){\color{white}%
	\frame{\includegraphics[width=0.4\linewidth, height=0.25\linewidth]{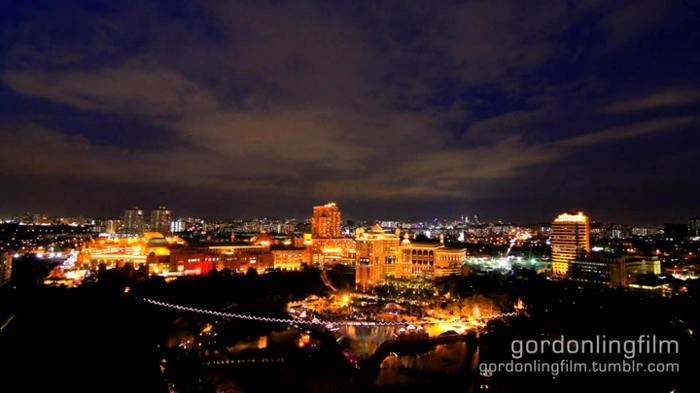}}}
\end{overpic}
\end{subfigure}
\begin{subfigure}[t]{0.19\linewidth}
\includegraphics[width=\linewidth, height=0.62\linewidth]{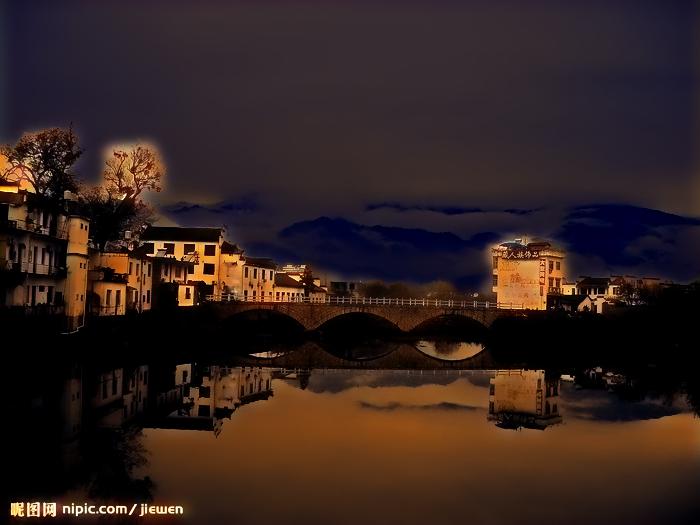}
\end{subfigure}
\begin{subfigure}[t]{0.19\linewidth}
\includegraphics[width=\linewidth, height=0.62\linewidth]{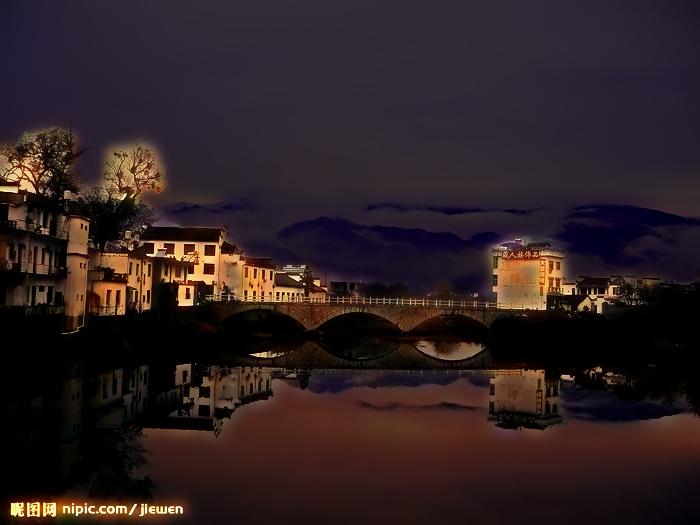}
\end{subfigure}
\begin{subfigure}[t]{0.19\linewidth}
\includegraphics[width=\linewidth, height=0.62\linewidth]{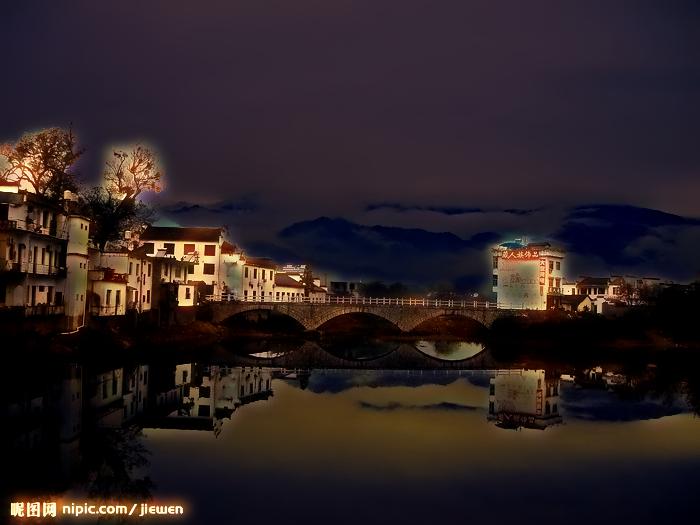}
\end{subfigure}
\begin{subfigure}[t]{0.19\linewidth}
\includegraphics[width=\linewidth, height=0.62\linewidth]{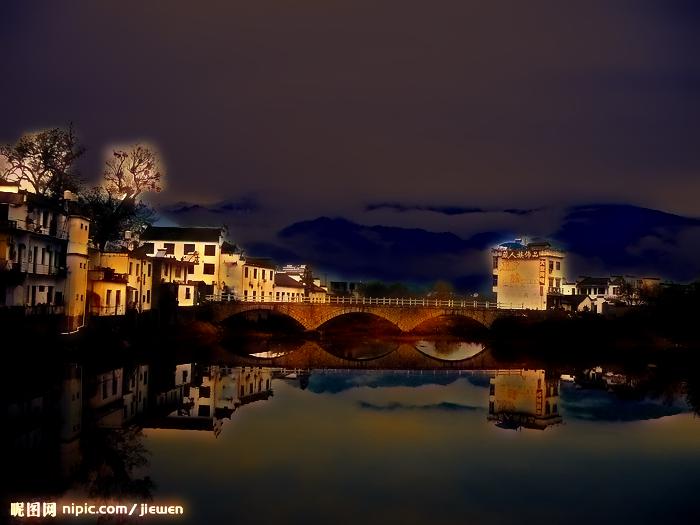}
\end{subfigure}
\vspace{1pt}

\begin{subfigure}[t]{0.19\linewidth}
\begin{overpic}[width=\linewidth, height=0.62\linewidth]{jpg/style/tar4.jpg}
\put(-2,38){\color{white}%
	\frame{\includegraphics[width=0.4\linewidth, height=0.25\linewidth]{jpg/content/in4.jpg}}}
\end{overpic}
\caption{\scalebox{1.15}{\tiny Input}}
\end{subfigure}
\begin{subfigure}[t]{0.19\linewidth}
\includegraphics[width=\linewidth, height=0.62\linewidth]{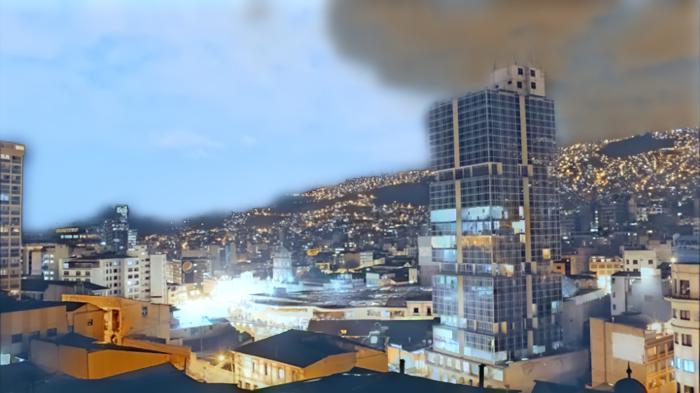}
\caption{\scalebox{1.15}{\tiny Full Training}}
\end{subfigure}
\begin{subfigure}[t]{0.19\linewidth}
\includegraphics[width=\linewidth, height=0.62\linewidth]{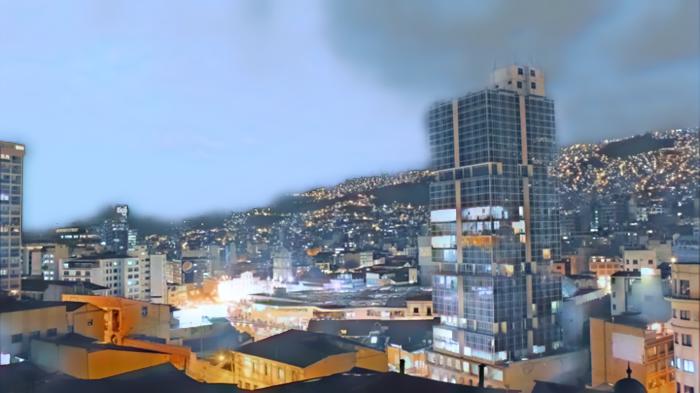}
\caption{\scalebox{1.15}{\tiny Partial Training}}
\end{subfigure}
\begin{subfigure}[t]{0.19\linewidth}
\includegraphics[width=\linewidth, height=0.62\linewidth]{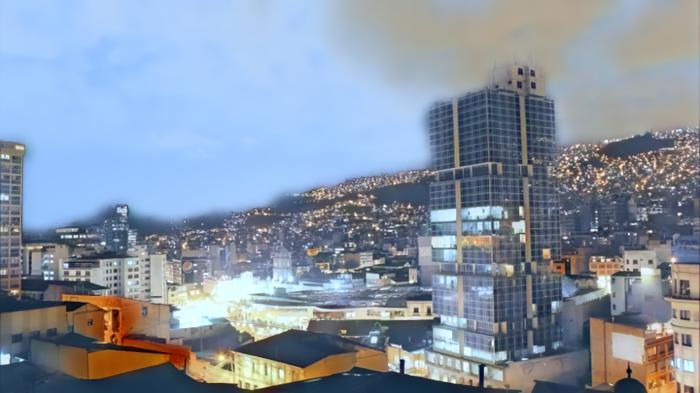}
\caption{\scalebox{1.15}{\tiny Partial Training}}
\end{subfigure}
\begin{subfigure}[t]{0.19\linewidth}
\includegraphics[width=\linewidth, height=0.62\linewidth]{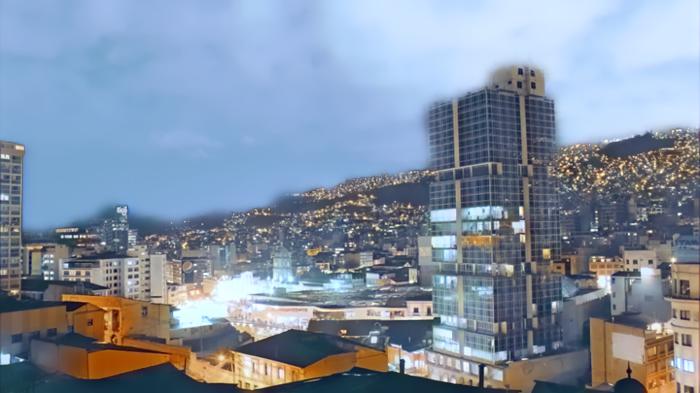}
\caption{\scalebox{1.15}{\tiny Partial Training}}
\end{subfigure}
\vspace{1pt}

\end{center}
   \caption{
Style networks are pre-trained for each image pair in the first row. Fixing the convolutional layers of these pre-trained models, we retrain the normalization parameters to transfer the styles between the photos in (a). The results are the two bottom rows of (c) to (e) for each pre-trained model. 
The results generated by randomly initialized and fully trained networks are presented in (b).
   }
\label{prefix}
\end{figure}

We mentioned in Section~\ref{Implemention Details} that our stylization networks share the same convolutional layers but have different instance normalization parameters, which control the styles. This design is due to Dumoulin~\etal\cite{dumoulin17}. 
We show below that it is sufficient to retrain the instance normalization parameters to adapt our networks to a new style, even though the convolutional layers are pre-trained using a single pair of images.
Given the extreme scarcity of training data, this property was not guaranteed, as the role of the convolutional filters and normalization parameters could have not been completely disentangled, letting the former still control part of the stylization. 

We trained a pair of networks on a first pair of images. We fix the convolutional layers and retrain both networks using another pair of images by optimizing only on the instance normalization parameters. Fig.~\ref{prefix}c-e present results obtained with this approach. The results are qualitatively comparable to those generated by 
randomly initialized and fully trained networks. 
Let us highlight nevertheless that this adaptation of the instance normalization parameters works better when the networks are pre-trained on two images of completely different styles (colorwise).

\section{Conclusion}
We designed a new method for effective photo stylization between two images that consists in training a pair of deep convnets with cycle- and self-consistency losses. Despite high-quality results on several examples, there is still room for improvement of our results. In particular, we should try to reduce the artifacts at the boundary of different semantic regions and reduce the overexposure that sometimes appears in small regions of the results. 
A direct extension of this work could be to train another network to predict the style parameters in $g_\theta(\cdot)$ directly from an image, while using the proposed loss, to be able to use arbitrary style at runtime.


\bibliographystyle{IEEEbib}
\bibliography{egbib}

\end{document}